\PassOptionsToPackage{table}{xcolor}
\documentclass[runningheads]{llncs}
\usepackage{accv}

\let\PointMFSuppSubsection\subsection
\let\PointMFSuppSubsubsection\subsubsection

\makeatletter
\renewcommand\subsection{\@startsection{subsection}{2}{\z@}%
                       {-12\p@ \@plus -4\p@ \@minus -4\p@}%
                       {7\p@ \@plus 4\p@ \@minus 4\p@}%
                       {\normalfont\normalsize\bfseries\boldmath
                        \rightskip=\z@ \@plus 8em\pretolerance=10000 }}
\renewcommand\subsubsection{\@startsection{subsubsection}{3}{\z@}%
                       {-10\p@ \@plus -4\p@ \@minus -4\p@}%
                       {-0.5em \@plus -0.22em \@minus -0.1em}%
                       {\normalfont\normalsize\bfseries\boldmath}}
\makeatother

\usepackage{accvabbrv}
\usepackage[final]{graphicx}
\usepackage{booktabs}
\usepackage{tabularx,array}
\newcolumntype{Y}{>{\centering\arraybackslash}X}
\usepackage[table]{xcolor}
\usepackage{wrapfig}
\usepackage{mathtools}
\usepackage{amssymb}
\usepackage{placeins}
\usepackage{caption}
\usepackage[most]{tcolorbox}
\usepackage{algpseudocode}
\usepackage{CJKutf8}

\definecolor{refblue}{RGB}{235,244,255}
\definecolor{ablationgray}{gray}{0.92}

\definecolor{algbg}{RGB}{245,245,245}
\definecolor{algborder}{RGB}{150,150,150}
\definecolor{algcomment}{RGB}{120,155,165}
\definecolor{algkw}{RGB}{180,70,120}

\newtcolorbox{algobox}{
  enhanced,
  colback=algbg,
  colframe=algborder,
  boxrule=0.4pt,
  arc=0pt,
  boxsep=0pt,
  left=6pt,right=6pt,top=5pt,bottom=5pt
}

\newcounter{algoboxctr}
\renewcommand{\thealgoboxctr}{\arabic{algoboxctr}}

\usepackage{hyperref}
\usepackage{orcidlink}

\begin{document}

% ============================================================
% MAIN PAPER
% ============================================================

\title{Point-MF: Stabilizing One-Step Mean Flows for Single-Image Point Cloud Reconstruction}

% TODO REVIEW: If the paper title is too long for the running head, you can set
% an abbreviated paper title here. If not, comment out.
\titlerunning{Point-MF}

% TODO FINAL: Replace with your author list. 
% Include the authors' OCRID for the camera-ready version, if at all possible.
\author{Yuta Baba\inst{1}\orcidlink{0009-0009-2729-3011} \and
Keiji Yanai\inst{1}\orcidlink{0000-0002-0431-183X} }

% TODO FINAL: Replace with an abbreviated list of authors.
\authorrunning{Y.~Baba and K.~Yanai}

\institute{
The University of Electro-Communications, Tokyo, Japan\\
\email{baba-y@mm.inf.uec.ac.jp, yanai@cs.uec.ac.jp}
}

\maketitle

\begin{abstract}
Single-image point cloud reconstruction requires recovering complete object-level geometry, including occluded regions, from a single RGB image.
Diffusion- and flow-based reconstructors can model this ambiguity, but their iterative sampling requires many network function evaluations, while naive one-step point-space updates often produce outliers and density imbalance.
We propose Point-MF, a one-step conditional Mean-Flow framework for geometry-only point cloud reconstruction.
Point-MF predicts an interval-averaged velocity field directly in point-cloud space and reconstructs the output with a single network function evaluation, without training a point-cloud autoencoder.
To stabilize large Mean-Flow jumps, we introduce Denoised Space Anchor (DSA), a set-distance auxiliary loss that anchors the denoised point set implied by the predicted velocity to the ground-truth geometry.
On all 13 ShapeNet-R2N2 categories and Pix3D, Point-MF achieves strong reconstruction quality under the aligned point-set protocol, improving average CD and EMD over the evaluated point-cloud reconstruction baselines.
It runs in 40.72 ms per sample on an RTX A4000, within the same latency order as the feedforward RGB2point baseline and substantially faster than iterative diffusion baselines such as PC$^2$ and BDM.
These results show that denoised-space geometric anchoring enables stable one-step Mean Flow directly in point-cloud space.
\end{abstract}

\section{Introduction}
\label{sec:intro}

Single-image point cloud reconstruction is an inherently ill-posed problem: a model must infer a complete 3D shape, including self-occluded and unobserved regions, from a single RGB observation. Since the same visible evidence can correspond to multiple plausible 3D completions, this task requires both image-conditioned shape inference and a mechanism for handling ambiguity. Fast reconstruction of sparse object-level geometry is useful for applications such as spatial understanding, robotic perception, AR/VR preprocessing, and lightweight downstream 3D analysis, where latency and memory cost are important practical constraints.

Point clouds support set-distance supervision and can serve as intermediate
representations for surface reconstruction or rendering~\cite{spar3d,ga,pointnerf,poco,sdfcon},
or as inputs to grasp prediction~\cite{graspnet}.
When only point geometry is needed, direct prediction avoids reconstructing
mesh connectivity that would be discarded after point sampling.
We therefore target fast, aligned geometry-only reconstruction rather than
textured asset generation.

Existing single-image 3D methods can be broadly grouped into direct regression methods, probabilistic diffusion/flow-based generative methods, and open-category image-to-3D models.
Direct regression methods~\cite{3dr2n2,pixel2mesh,rgb2point,sf3d,lrm,shapeclip,crm} predict a 3D representation from an input image in a single forward pass and are therefore attractive for fast inference.
Among point-cloud reconstruction methods, RGB2point~\cite{rgb2point} is a strong practical baseline.
However, deterministic prediction makes it difficult to represent multiple plausible shape completions for occluded regions.

\begin{figure}[tb]
\makebox[\linewidth][c]{
\includegraphics[width=\linewidth]{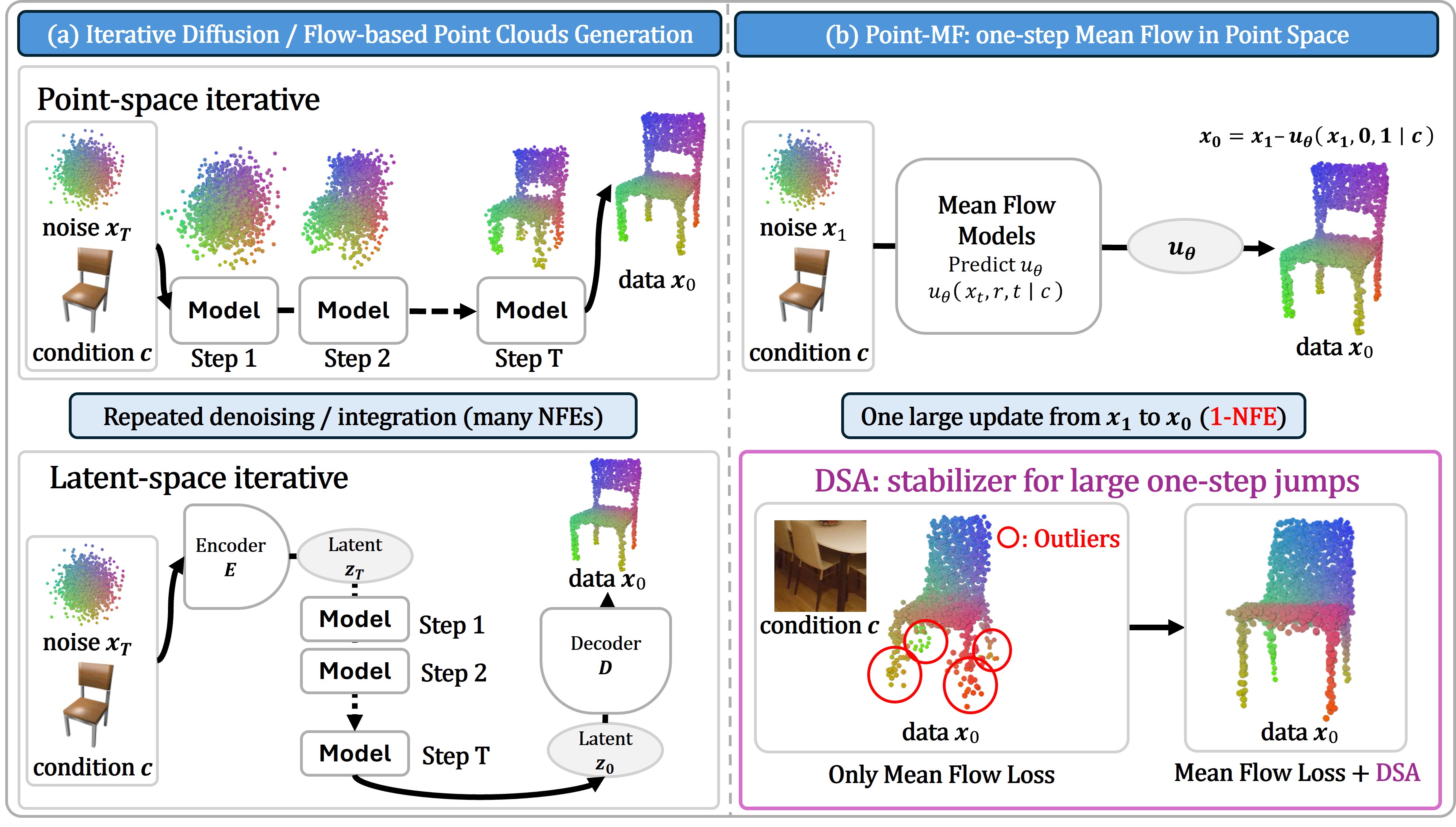}
}
\caption{
Conceptual comparison between Point-MF and iterative point-cloud generation.
(a) Many existing diffusion/flow-based point-cloud generators obtain reconstructions through iterative denoising or numerical integration, requiring many network function evaluations (NFEs).
(b) Point-MF directly predicts an interval-averaged velocity field in point-cloud space and performs a one-step update from noise to the reconstructed point cloud.
DSA regularizes the induced data-side estimate with a set-distance loss, suppressing outliers and density artifacts caused by one-step updates.
}
\label{fig:intro}
\vspace{-8mm}
\end{figure}

In contrast, diffusion/flow-based methods~\cite{pc2,bdm,pointe,lion,dit3d,pointinfinity,diffusionsddf,shap3,ln3diff,chen2023single} can probabilistically model shape ambiguity, but they typically require many NFEs due to iterative denoising or numerical integration.
More recently, open-category image-to-3D models and asset generators, including the TRELLIS series~\cite{trellis,trellis2} and related methods~\cite{spar3d,sf3d,sparseflex,lattice,ultrashape}, have substantially advanced high-resolution 3D generation.
These models are important reference points, but many of them target 3D asset generation with meshes, structured 3D latent representations, 3D Gaussians, or appearance and material information.
This differs from the aligned geometry-only point cloud reconstruction setting studied in this work.

Mean Flow~\cite{meanflow}, derived from Flow Matching~\cite{flowmatching}, enables large updates with few sampling steps by learning an interval-averaged velocity field between two time points.
However, directly applying this idea to unordered point-cloud space is non-trivial.
An $N$-point cloud lies in a high-dimensional set space, and even small velocity errors in a one-step update can appear as outliers, density imbalance, or deviations from the underlying shape manifold.
Therefore, the challenge is to reduce NFEs while stabilizing the clean
geometry (a data-side estimate at $t=0$) induced by a large Mean Flow update.

To address this challenge, we propose \textbf{Point-MF}, a one-step conditional Mean Flow framework for single-image point cloud reconstruction (Fig.~\ref{fig:intro}).
Point-MF predicts the mean velocity field directly in point-cloud space without using a point-cloud autoencoder, and reconstructs the output with a single network function evaluation.
The central component is \textbf{Denoised Space Anchor} (DSA).
DSA anchors the clean point-cloud estimate $x_\theta$ induced by the predicted velocity to the ground-truth geometry using a set-distance loss, thereby suppressing outliers and density artifacts caused by large one-step updates.

We evaluate Point-MF on ShapeNet-R2N2 and Pix3D in terms of both shape quality and efficiency.
On ShapeNet-R2N2, we report results on all 13 categories and compare Point-MF with feedforward methods and iterative diffusion-based methods.
For recent image-to-3D models, we provide representative qualitative comparisons in the main paper, and report quantitative reference evaluations involving point-cloud conversion in the supplementary material.

Our contributions are summarized as follows:
\begin{itemize}
\item We formulate single-image point cloud reconstruction as one-step conditional Mean Flow directly in point-cloud space, and propose Point-MF, a geometry-only reconstruction framework that does not require a point-cloud autoencoder.

\item We introduce Denoised Space Anchor (DSA), which regularizes the clean point-cloud estimate induced by the predicted mean velocity using a set-distance loss, reducing outliers and density artifacts caused by one-step updates.

\item We validate Point-MF through comprehensive experiments on all 13 ShapeNet-R2N2 categories and Pix3D, including reconstruction quality, inference efficiency, memory usage, and ablation studies.
\end{itemize}

\section{Related Work}

\subsubsection{3D Reconstruction from Images.}
Image-based 3D reconstruction has explored a wide range of 3D representations,
including voxels, meshes, point clouds, implicit functions, and Gaussian Splatting
~\cite{3dr2n2,genre,pixel2mesh,atlas,occ,deepsdf,3dgs,gsd}.
For single-image reconstruction, prior work has proposed voxel-based methods
such as 3D-R2N2~\cite{3dr2n2}, mesh-based methods such as Pixel2Mesh and
AtlasNet~\cite{pixel2mesh,atlasnet}, and implicit-function-based methods such as
Occupancy Networks and DeepSDF~\cite{occ,deepsdf}.
In contrast, point clouds provide a lightweight geometry representation and can be
directly supervised with set distances, making them suitable for geometry-only
single-image reconstruction.
Representative point-cloud reconstruction methods include reprojection-based
approaches~\cite{ulsp} and RGB2point~\cite{rgb2point}, which performs fast direct
point-cloud generation.

More recently, probabilistic generation of point clouds and 3D shapes has been
widely studied.
Point cloud diffusion~\cite{pc2gen}, DiT-3D~\cite{dit3d}, Point-E~\cite{pointe},
PC$^2$~\cite{pc2}, and BDM~\cite{bdm} are representative examples that extend
diffusion models to 3D generation or single-image reconstruction.
These methods can probabilistically model ambiguous 3D completion, but they require
many iterative updates at inference time, resulting in large NFEs.
In contrast, LION~\cite{lion} is a representative latent-space diffusion method
that uses a hierarchical point-cloud VAE and performs diffusion generation in a latent
space consisting of a global shape latent and a point-structured latent.
While LION achieves stronger generation performance than DDMs operating directly
in point-cloud space, it depends on the training and reconstruction quality of a
point-cloud autoencoder and still requires iterative sampling during generation.
In contrast, our method learns one-step conditional Mean Flow directly in
point-cloud space without using a point-cloud autoencoder.

In addition, open-category image-to-3D models and 3D asset generation methods,
such as LRM~\cite{lrm}, the TRELLIS series~\cite{trellis,trellis2},
SPAR3D~\cite{spar3d}, SAM3D~\cite{sam3d},
ZeroShape/ZeroShape-W~\cite{zeroshape,zerow}, and SparseFlex~\cite{sparseflex},
have made substantial progress in high-resolution and open-category 3D generation.
However, many of these methods target 3D asset generation with meshes, structured
3D latent representations, dense geometry, or appearance information.
This differs from the aligned geometry-only point cloud reconstruction setting studied
in this work in terms of output representation, training data, and evaluation protocol.
Therefore, we treat these methods not as primary in-domain competitors, but as
reference methods under a point-sampled geometry evaluation protocol.

\subsubsection{Diffusion and Flow-based Models.}
Diffusion and flow-based generative models learn a mapping from a prior distribution
to a data distribution as a continuous-time generative trajectory
~\cite{deepunsup,ddpm,edm,scoresde,scorematch,flowmatching,recflow}.
While these methods enable high-quality generation, they often require iterative
numerical integration at inference time, which leads to high computational cost.
Recent work has also studied regularizing the clean output $x_0$ estimated from
intermediate states along the generative trajectory to improve few-step generation
quality~\cite{cm,icm,bring,irf}.
In this work, we extend this clean-output supervision principle to point clouds by
anchoring the induced denoised point set to the ground-truth geometry using a
set-distance loss.

\subsubsection{Few-step Generative Models.}
Standard diffusion and flow-based models assume multi-step numerical integration,
whereas recent methods aim at one-step or few-step generation by directly learning
large temporal updates~\cite{cm,icm,cm_easy,ct_consis,ctm,shortcut,imm,meanflow}.
Consistency Models, CTM, Shortcut Models, and IMM accelerate generation through
direct mappings to terminal samples, trajectory learning between arbitrary time pairs,
or moment matching~\cite{ctm,shortcut,imm}.
Mean Flow~\cite{meanflow} enables low-NFE generation by learning the mean velocity
field between two time points, and several follow-up methods have improved its
stability and sampling behavior~\cite{imf,alpha,dmf}.
However, as shown by Pixel Mean Flows~\cite{pmf}, naive $u$-prediction can become
unstable in high-dimensional latent-free spaces.
A similar issue arises in point-cloud space: errors in a one-step velocity prediction can
manifest as outliers, density imbalance, or shape deviations.
We therefore stabilize the induced denoised estimate geometrically using DSA.

\section{Preliminaries}

\subsubsection{Flow Matching.}
Flow Matching (FM)~\cite{flowmatching} learns a velocity field along a transport path
between a data distribution and a prior distribution.
Given a data sample $x\sim p_{\mathrm{data}}$ and noise
$\epsilon\sim p_{\mathrm{prior}}$, we use the linear interpolation
$x_t=(1-t)x+t\epsilon$, whose path-wise velocity is $\epsilon-x$.
FM regresses the marginal velocity field induced by this path,
$v(x_t,t)=\mathbb{E}[\epsilon-x\mid x_t]$.
Under our time convention, $t=0$ corresponds to the data side and $t=1$ to the noise side.
At inference time, samples are generated by integrating the learned ODE backward from
$t=1$ to $t=0$.

\subsubsection{Mean Flow.}
Mean Flow (MF)~\cite{meanflow} enables low-NFE generation by directly learning an interval-averaged velocity field between two time points, rather than the instantaneous velocity field used in Flow Matching.
Under our time convention, $t=0$ corresponds to the data side and $t=1$ to the noise side.
For $0 \le r < t \le 1$, the mean velocity field for moving from the state $x_t$ at time $t$ to the state $x_r$ at time $r$ is defined as
\begin{align}
u^\star(x_t,r,t)
\coloneqq
\frac{1}{t-r}
\int_{r}^{t}
v^\star(x_\tau,\tau)\,d\tau .
\label{eq:mf_mean_velocity}
\end{align}
Using this mean velocity field, the update over the interval $[r,t]$ can be written as
\begin{align}
x_r = x_t - (t-r)u^\star(x_t,r,t).
\label{eq:mf_jump}
\end{align}
Therefore, if $u^\star$ can be directly approximated by a neural network,
a large temporal jump can be performed without many numerical integration steps.

In practice, instead of directly evaluating the integral in Eq.~\eqref{eq:mf_mean_velocity},
MF uses a self-consistent regression target derived from the Mean Flow identity.
In this work, we introduce an image condition $c$ and train a conditional MF model
$u_\theta(x_t,r,t\mid c)$ with classifier-free condition dropout.
The detailed derivation of Classifier-Free Guidance (CFG) -guided Mean Flow and the target computation using Jacobian--vector product (JVP) are provided in Appendix~\ref{sec:supp_cfg_mean_flow}.
In the main paper, we use the following objective:
\begin{align}
\mathcal{L}_{\mathrm{MF\text{-}CFG}}(\theta)
=
\mathbb{E}
\left[
\left\|
u_\theta(x_t,r,t\mid c)
-
\mathrm{sg}
\left(
u_{\mathrm{tgt}}^{\mathrm{cfg}}
\right)
\right\|_2^2
\right],
\label{eq:cfg_loss}
\end{align}
where $u_{\mathrm{tgt}}^{\mathrm{cfg}}$ denotes the  CFG-guided regression target based on the Mean Flow identity, and $\mathrm{sg}(\cdot)$ denotes stop-gradient.

At inference time, we fix $(r,t)=(0,1)$ and generate a point cloud from noise $x_1=\epsilon$ by
\begin{align}
x_0
=
\epsilon
-
u_\theta(\epsilon,0,1\mid c),
\label{eq:mf_inference}
\end{align}
using a single network function evaluation.

\section{Method}
\label{sec:method}

\begin{figure}[tb]
\makebox[\linewidth][c]{
\includegraphics[width=\linewidth]{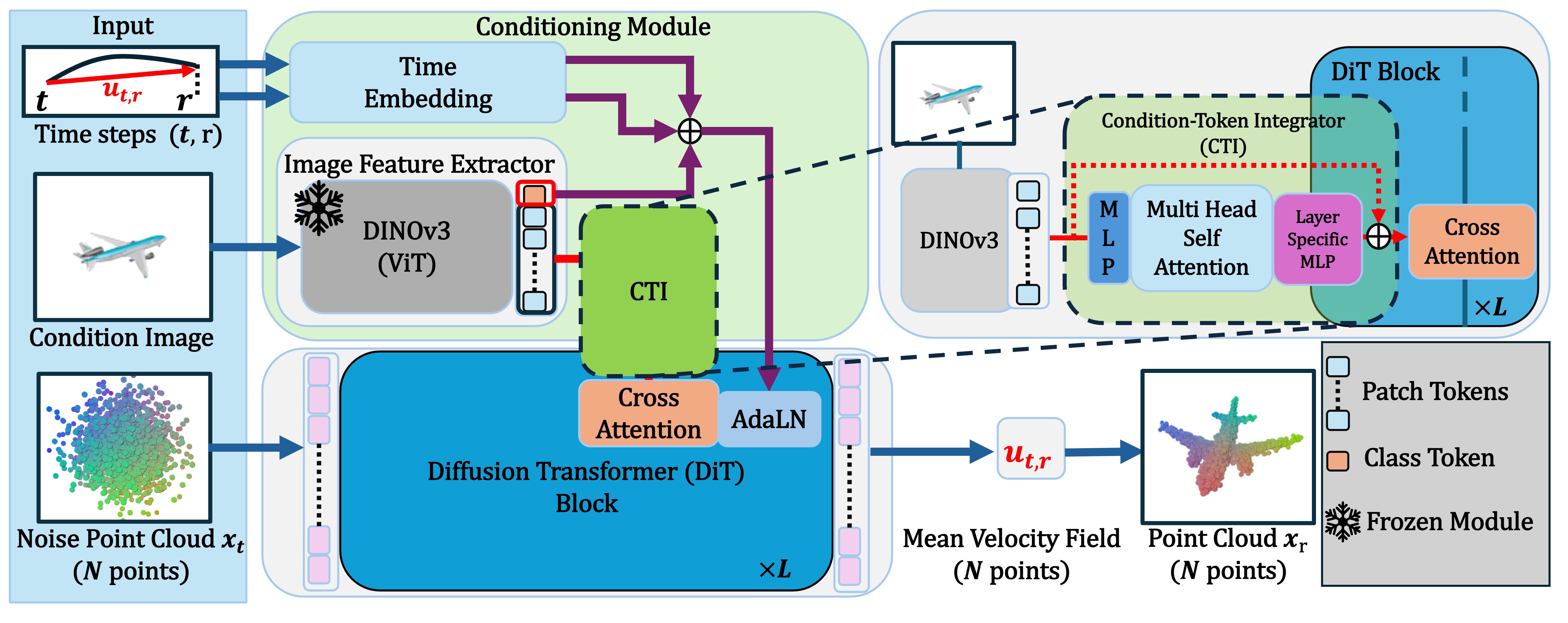}
}
\caption{
Overview of Point-MF.
Global features from a frozen DINOv3 encoder are combined with the embeddings of time $t$ and interval length $dt=t-r$,
and are used for AdaLN-Zero modulation.
In CTI, patch tokens are integrated once by shared processing, and each DiT block forms its Cross-Attention context through a layer-specific MLP.
The point-cloud DiT predicts the mean velocity field $u_\theta(x_t,r,t\mid c)$ from the noisy point cloud $x_t$,
and generates the point cloud through the Mean Flow update $x_r=x_t-(t-r)u_\theta(x_t,r,t\mid c)$.
}
\label{fig:arch}
\vspace{-5mm}
\end{figure}

This section describes Point-MF, a conditional Mean Flow framework for single-image point cloud reconstruction.
We first describe the network architecture, focusing on image conditioning, interval/time conditioning, and the proposed Condition-Token Integrator (CTI).
An overview is shown in Fig.~\ref{fig:arch}.
We then present the loss design for stabilizing one-step Mean Flow generation and improving geometric consistency in data space.

\subsection{Architecture}

\subsubsection{Image Conditioning with DINOv3.}
In single-image point cloud reconstruction, the model must infer complete 3D geometry from partial visual evidence.
We use a frozen DINOv3 encoder~\cite{DINO3} to extract image features from the input image.
The encoder outputs a global image feature $\mathbf{z}_{\mathrm{img}}\in\mathbb{R}^{D}$ corresponding to the class token,
and patch-sequence features $Z_{\mathrm{ctx}}\in\mathbb{R}^{M\times D}$.
The global feature is projected to the hidden dimension of the DiT and used as $e_{\mathrm{img}}\in\mathbb{R}^{H}$ for global modulation.
The patch features are fed into the Condition-Token Integrator (CTI), described in Sec.~\ref{sec:cti}, and are transformed into layer-wise context tokens for Cross-Attention.
The DINOv3 encoder is frozen, while the projection layers, CTI, and point-cloud generator are trained for the reconstruction objective.
\subsubsection{Mean-Flow-Compatible Point-Cloud DiT.}
The point-cloud generator is implemented as a Transformer composed of $L$ Diffusion Transformer (DiT) blocks.
Since a point cloud is an unordered set, we follow prior point-cloud diffusion models~\cite{pointe} and do not use point-index-dependent positional embeddings,
thereby preserving permutation equivariance with respect to the input point order.
The DiT backbone is based on the sparse point-cloud generation Transformer of Lan \etal~\cite{ga}.
Unlike standard diffusion models, our network predicts the interval-averaged velocity field $u_\theta(x_t,r,t\mid c)$ rather than an instantaneous velocity.
Therefore, in addition to the current time $t$, we condition the model on the interval length $dt=t-r$.

Following DiT~\cite{DiT}, we map $t$ and $dt$ into
$e_t(t), e_{dt}(dt)\in\mathbb{R}^{H}$ using sinusoidal embeddings followed by two-layer MLPs.
We then add the global image embedding $e_{\mathrm{img}}$ to obtain the conditioning vector
$\mathbf{I}=e_t(t)+e_{dt}(dt)+e_{\mathrm{img}}$.
This vector is used to generate the AdaLN-Zero modulation parameters in each DiT block.
Since Mean Flow estimates an average velocity over an interval rather than an instantaneous velocity at time $t$,
explicitly conditioning on $dt$ enables the model to handle different temporal jumps.
The detailed block-level architecture is provided in Appendix~\ref{sec:supp_impl}.

\subsubsection{Condition-Token Integrator (CTI).}
\label{sec:cti}

Patch tokens from a frozen DINOv3 encoder contain strong semantic information, but they are not directly optimized as Cross-Attention conditions for point-cloud generation.
We therefore introduce the \emph{Condition-Token Integrator (CTI)}, which transforms image patch tokens into layer-wise context tokens for the point-cloud DiT.
CTI consists of a token integration stage applied once on the condition branch and a lightweight layer-specific injection stage inside each DiT block.

Let $Z_{\mathrm{ctx}}\in\mathbb{R}^{M\times D}$ denote the patch features extracted by DINOv3.
We first project $Z_{\mathrm{ctx}}$ with an input projection $\phi_{\mathrm{in}}$ to obtain
$Z_{\mathrm{base}}=\phi_{\mathrm{in}}(Z_{\mathrm{ctx}})\in\mathbb{R}^{M\times H}$.
We then apply MHSA once to the projected image tokens and obtain integrated features
$A_{\mathrm{CTI}}=\mathrm{MHSA}(\phi_{\mathrm{MHSA}}(Z_{\mathrm{base}}))$.
In the $\ell$-th DiT block, this integrated feature is projected by a layer-specific MLP $\psi_\ell$ and added to the base image tokens:
\begin{align}
Z_{\mathrm{CTI}}^{(\ell)}
=
Z_{\mathrm{base}} + \psi_{\ell}(A_{\mathrm{CTI}}).
\label{eq:cti}
\end{align}
The resulting $Z_{\mathrm{CTI}}^{(\ell)}$ is used as the Cross-Attention context in the $\ell$-th DiT block.

In this design, token interactions are computed only once on the condition branch, while each DiT block applies only a lightweight layer-specific projection.
Thus, CTI provides a task-specific interface between frozen DINOv3 features and the point-cloud DiT while keeping the generator compact.

\subsection{Training Objective}
\label{sec:loss}

\subsubsection{Denoised Space Anchor.}

\begin{figure}[tb]
  \centering
  \includegraphics[width=0.8\linewidth]{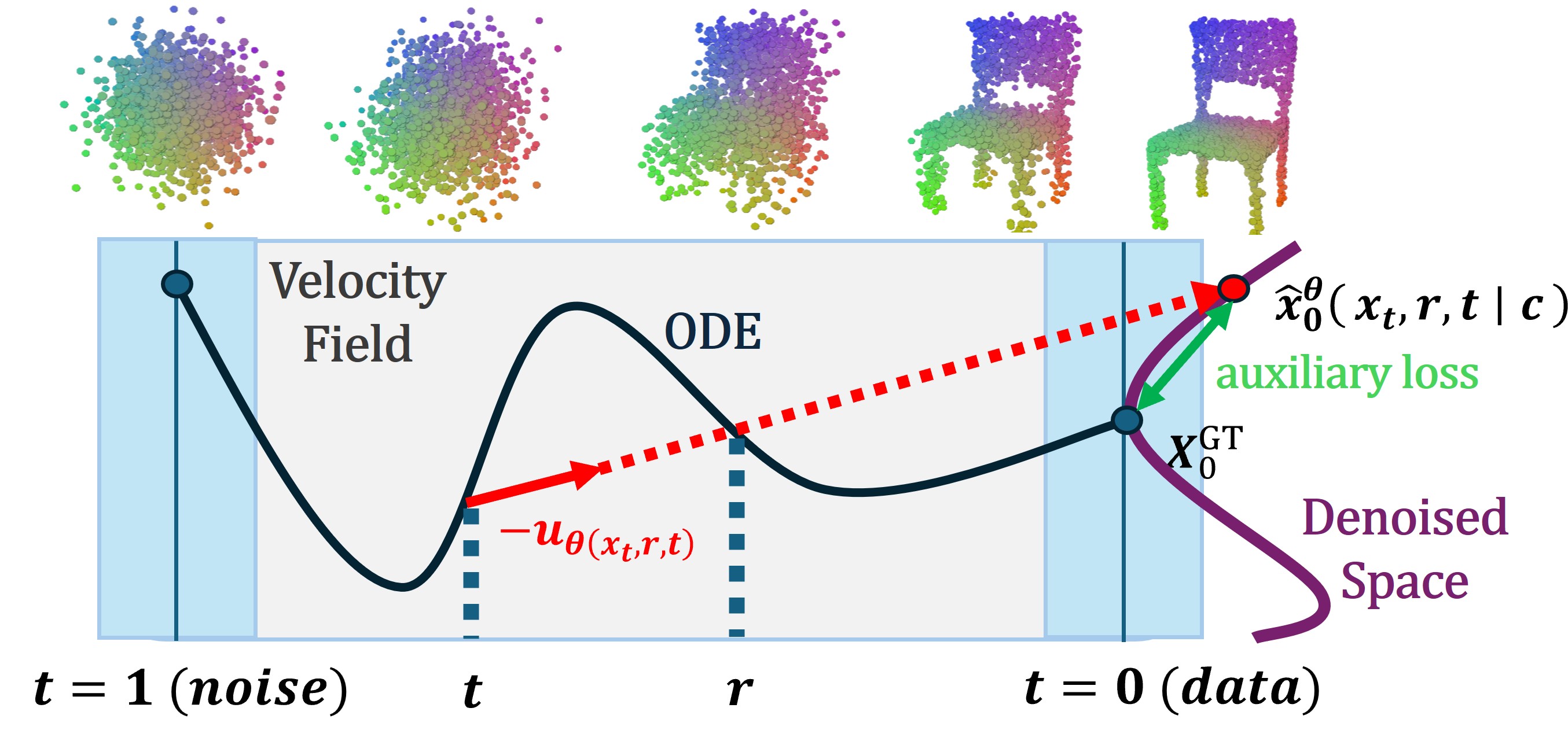}
\caption{
Conceptual illustration of Denoised Space Anchor (DSA).
Point-MF extrapolates the predicted mean velocity field $u_\theta(x_t,r,t\mid c)$ toward the data time $t=0$ to obtain a denoised-space estimate $\hat{x}_0^\theta$.
DSA anchors this estimated point cloud to the ground-truth point cloud $x_0^{\mathrm{GT}}$ using a set-distance loss,
thereby suppressing outliers and density imbalance caused by one-step Mean Flow updates.
}
  \label{fig:dsa}
  \vspace{-5mm}
\end{figure}

Mean Flow enables low-NFE generation by predicting an interval-averaged velocity
field for a large temporal jump. In point-cloud space, however, small velocity errors can
directly lead to outliers, surface misalignment, or density imbalance in the reconstructed
shape. This is particularly problematic in one-step generation, where there are no later
correction steps.

To address this issue, we introduce \emph{Denoised Space Anchor (DSA; Fig.~\ref{fig:dsa})}, an auxiliary
loss that regularizes the clean point cloud implied by the predicted mean velocity. Given
a sampled state $x_t$ and a predicted mean velocity $u_\theta(x_t,r,t\mid c)$, we form
a denoised-space estimate by extrapolating the prediction to the clean endpoint:
\begin{align}
\hat{x}_0^\theta(x_t,r,t\mid c)
=
x_t
-
t\,u_\theta(x_t,r,t\mid c).
\label{eq:x0_pred_dsa}
\end{align}
When $r=0$, Eq.~\eqref{eq:x0_pred_dsa} has the Mean-Flow update form.
For general $(r,t)$, extrapolation to $t=0$ is exact for the constant
velocity of a sample-wise linear path, but not generally for the
marginal field~\cite{meanflow}; we use it as an auxiliary geometric estimate.

Since point clouds are unordered sets, a point-wise loss is not appropriate unless a
correspondence is given. We therefore use Adaptive Probabilistic Matching Loss
(APML)~\cite{APML} as the set distance:
\begin{align}
\mathcal{L}_{\mathrm{DSA}}
=
\mathcal{L}_{\mathrm{APML}}
\left(
\hat{x}_0^\theta(x_t,r,t\mid c),
x_0^{\mathrm{GT}}
\right).
\label{eq:dsa_loss}
\end{align}
APML provides soft matching between the predicted and ground-truth point sets, making
it better suited to the denoised-space estimate than a direct point-wise regression loss.
In Sec.~\ref{sec:ablations}, we further compare APML with Chamfer Distance and MSE
as the set distance used in DSA.

DSA is used as an auxiliary constraint and does not replace the Mean-Flow velocity
supervision. It instead pulls the clean point cloud induced by the predicted velocity field
toward geometrically plausible shapes. To balance the strength of this constraint across
time, we use the time-dependent weight
\begin{align}
\lambda(t)
=
\mathbf{1}_{t\neq r}
\frac{\lambda_{\mathrm{base}}}{\max(t,\tau)}.
\label{eq:dsa_time_weight}
\end{align}
We set $\tau=10^{-5}$ to bound the weight near $t=0$.
The factor $1/\max(t,\tau)$ compensates for the factor $t$ in
Eq.~\eqref{eq:x0_pred_dsa}, while $\mathbf{1}_{t\ne r}$ disables
DSA for the local Flow-Matching-like branch ($t=r$).

Finally, to absorb the scale difference between the Mean-Flow objective and the set
distance, we use a mini-batch-based scaling coefficient:
\begin{align}
s
=
\frac{
\mathbb{E}[\mathcal{L}_{\mathrm{MF\text{-}CFG}}]_{\mathrm{detach}}
}{
\mathbb{E}[\mathcal{L}_{\mathrm{DSA}}]_{\mathrm{detach}}
+
\delta
}.
\label{eq:dsa_scale}
\end{align}
The stop-gradient on the batch averages prevents the scaling coefficient itself from
becoming an optimization target. The small constant $\delta$ is used only for numerical
stability; we set $\delta=10^{-8}$ in all experiments.

The final training objective is
\begin{align}
\mathcal{L}_{\mathrm{total}}
=
\mathcal{L}_{\mathrm{MF\text{-}CFG}}
+
s\,\lambda(t)\,\mathcal{L}_{\mathrm{DSA}}.
\label{eq:total_loss_dsa}
\end{align}
For $t\ne r$, the explicit DSA gradient-scale factor is
$s\lambda_{\mathrm{base}}\min(1,t/\tau)$,
so the clipping affects this factor only for $t<\tau$.
In our experiments, we set $\lambda_{\mathrm{base}}=0.5$, which provides the best
EMD-oriented trade-off in the DSA weight ablation.

\section{Experiments}
We evaluate Point-MF in terms of both reconstruction quality and inference efficiency.
Since our goal is to generate geometrically consistent point clouds with low-NFE sampling,
we report not only CD, EMD, and F-Score, but also inference time and peak memory usage.

\subsubsection{Datasets.}
We evaluate Point-MF on ShapeNet-R2N2~\cite{shapenet} and Pix3D~\cite{pix3d}.
For ShapeNet-R2N2, we follow the 13-category split used in prior work
~\cite{3dr2n2,selfsup}, sample target point clouds from each 3D model,
and normalize them to be centered at the origin.
For Pix3D, we follow the split of Xu \etal~\cite{bdm}.
Pix3D is used only as a test-only benchmark to evaluate real-image generalization,
and is not used for training.

\begin{table}[tb]
  \centering
  \caption{
  Comparison of CD and EMD on ShapeNet-R2N2.
NFE denotes inference-time network evaluations; BDM uses $256+20=276$ NFEs
including its prior model.
``--'' denotes unavailable or incompatible comparisons.
ViT-B uses a standard ImageNet-pretrained encoder~\cite{VisionTransformer}.
For FCM~\cite{fcm}, we report both 1-view and 2-view settings.
  }
  \vskip -2mm
  \label{tab:compare_cd_emd}
  \scriptsize
  \setlength{\tabcolsep}{2.7pt}
  \begin{tabularx}{\linewidth}{l c *{8}{Y}}
    \toprule
     & & \multicolumn{4}{c}{CD$\times 100$ $\downarrow$}
     & \multicolumn{4}{c}{EMD$\times 100$ $\downarrow$} \\
    \cmidrule(lr){3-6} \cmidrule(lr){7-10}
    Method & NFE & Car & Chair & Air & Avg. & Car & Chair & Air & Avg. \\
    \midrule
    Self-Sup.~\cite{selfsup}
    & 1 & 5.48 & 10.91 & 7.11 & 7.85
    & 4.95 & 14.93 & 11.07 & 9.99 \\

    DIFFER~\cite{differ}
    & 1 & 6.35 & 9.78 & 5.67 & 7.47
    & 6.03 & 16.21 & 9.90 & 10.65 \\

    ULSP~\cite{ulsp}
    & 1 & 5.40 & 9.72 & 5.91 & 7.11
    & 4.78 & 10.18 & 7.66 & 7.41 \\

    RGB2point~\cite{rgb2point}
    & 1 & \textbf{4.22} & 5.43 & \textbf{2.70} & \underline{4.33}
    & 5.63 & 9.53 & 6.86 & 7.34 \\

    \midrule
    PC$^2$~\cite{pc2}
    & 256 & 4.95 & 5.60 & 3.37 & 4.84
    & 4.69 & 5.36 & 4.27 & 4.85 \\

    BDM~\cite{bdm}
    & 276 & 5.12 & \underline{5.02} & 3.12 & 4.64
    & \underline{4.25} & \underline{4.53} & \underline{3.92} & \underline{4.28} \\

    LION~\cite{lion}
    & 1000 & 5.86 & 12.53 & 5.95 & 8.35
    & 6.54 & 13.73 & 8.08 & 9.54 \\

    FCM (1-View)~\cite{fcm}
    & 256 & 4.93 & -- & 3.52 & --
    & 5.57 & -- & 5.39 & -- \\

    FCM (2-View)~\cite{fcm}
    & 256 & \underline{4.64} & -- & 2.91 & --
    & 5.51 & -- & 5.17 & -- \\
    \midrule
    Ours
    & 1 & \textbf{4.22} & \textbf{4.29} & \underline{2.76} & \textbf{3.92}
    & \textbf{3.82} & \textbf{4.29} & \textbf{3.04} & \textbf{3.82} \\

    \rowcolor{gray!15}
    w/o CTI
    & 1 & 4.58 & 4.29 & 3.01 & 4.13
    & 4.02 & 4.62 & 3.54 & 4.14 \\

    \rowcolor{gray!15}
    w/o DSA
    & 1 & 5.21 & 6.16 & 3.71 & 5.23
    & 4.89 & 6.08 & 4.12 & 5.16 \\
    \rowcolor{gray!15}
    \rowcolor{gray!15}
    DINOv3 $\rightarrow$ ViT
    & 1 & 4.12 & 4.45 & 2.72 & 3.93
    & 4.02 & 4.31 & 3.12 & 3.93 \\

    \bottomrule
  \end{tabularx}
  \vspace{-6mm}
\end{table}
\subsubsection{Metrics and Evaluation Protocol.}
\label{sec:metrics}
We evaluate the reconstruction quality between the generated point cloud $\hat X$
and the ground-truth point cloud $X$ using L1 Chamfer Distance (CD),
Earth Mover's Distance (EMD), and F-Score.
CD and EMD are used as complementary metrics for measuring geometric accuracy and
global point-correspondence consistency.
For CD and EMD, we unify the output resolution to 1024 points on ShapeNet-R2N2
and 2048 points on Pix3D.
F-Score is computed with a threshold of 1\% of the longest side of the bounding box.
Avg.\ is the instance-weighted mean over listed categories
(ShapeNet-R2N2 counts: Table~\ref{tab:supp_shapenet_cd};
Pix3D images: chair/sofa/table = 768/390/374).
We reimplemented RGB2point.
\begin{table}[b]
  \vspace{-3mm}
  \centering
  \vskip -2mm
  \caption{
F-Score comparison on ShapeNet-R2N2 with a threshold of 1\%.
$\checkmark$ indicates evaluation after RepKPU upsampling from 2048 to 8192 points.
}
  \vskip -3mm
  \label{tab:compare_fs}
  \scriptsize
  \setlength{\tabcolsep}{1.8pt}
  \begin{tabular*}{\linewidth}{@{\extracolsep{\fill}}lccccccc}
  \toprule
  Category
  & {\scriptsize LegoFormer~\cite{lego}}
  & {\scriptsize PC$^2$~\cite{pc2}}
  & {\scriptsize BDM~\cite{bdm}}
  & {\scriptsize LION~\cite{lion}}
  & {\scriptsize RGB2point~\cite{rgb2point}}
  & {\scriptsize Ours}
  & {\scriptsize Ours} \\
  \midrule
  Upsample
  &
  & $\checkmark$
  & $\checkmark$
  & --
  & --
  & --
  & $\checkmark$ \\
  \midrule
    airplane    & 0.215 & 0.473 & 0.512 & 0.560 & 0.583 & \underline{0.591} & \textbf{0.603} \\
    car         & 0.220 & 0.301 & 0.311 & 0.272 & \underline{0.339} & 0.338 & \textbf{0.342} \\
    chair       & 0.217 & 0.202 & 0.238 & 0.145 & 0.195 & \underline{0.243} & \textbf{0.252} \\
    sofa        & \textbf{0.260} & 0.177 & 0.211 & --    & 0.194 & 0.238 & \underline{0.240} \\
    table       & 0.305 & 0.238 & 0.268 & --    & 0.268 & \underline{0.309} & \textbf{0.316} \\
    \midrule
    Average     & 0.247 & 0.271 & 0.299 & --    & 0.304 & \underline{0.332} & \textbf{0.339} \\
    \bottomrule
  \end{tabular*}
  \vspace{-5mm}
\end{table}

For F-Score, checkmarks in Table~\ref{tab:compare_fs} indicate
RepKPU~\cite{kpu} upsampling from 2048 to 8192 points;
unmarked Point-MF results use direct 8192-point generation without retraining.
This is enabled by the absence of index-dependent positional embeddings
and does not imply resolution-invariant performance.
\begin{table}[tb]
  \centering
  \vskip 4mm
  \caption{Comparison of CD and EMD on the Pix3D dataset.}
  \vskip -3mm
  \label{tab:compare_cd_emd_pix3d}
  \scriptsize
  \setlength{\tabcolsep}{2.6pt}
  \begin{tabularx}{\linewidth}{lc*{8}{Y}}
    \toprule
     & & \multicolumn{4}{c}{CD$\times 100$ $\downarrow$}
       & \multicolumn{4}{c}{EMD$\times 100$ $\downarrow$} \\
    \cmidrule(lr){3-6} \cmidrule(lr){7-10}
    Method & NFE & Chair & Sofa & Table & Avg. & Chair & Sofa & Table & Avg. \\
    \midrule
    ZeroShape~\cite{zeroshape}
    & 1 & 8.66 & 7.12 & 8.40 & 8.20 & 9.12 & 7.54 & 10.10 & 8.96 \\

    PC$^2$~\cite{pc2}
    & 256 & 7.07 & 5.87 & 11.39 & 7.82 & 7.13 & 6.14 & 10.72 & 7.75 \\

    BDM~\cite{bdm}
    & 276 & \underline{6.62} & 5.55 & \underline{8.35} & \underline{6.77}
    & \underline{7.05} & \textbf{4.98} & \underline{8.79} & \underline{6.95} \\

    RGB2point~\cite{rgb2point}
    & 1 & \underline{6.62} & \underline{5.40} & 9.70 & 7.06
    & 9.34 & 7.11 & 12.69 & 9.59 \\

    \midrule
    Ours
    & 1 & \textbf{6.48} & \textbf{5.02} & \textbf{8.20} & \textbf{6.53}
    & \textbf{6.74} & \underline{5.06} & \textbf{7.94} & \textbf{6.61} \\

    \rowcolor{gray!15}
    w/o CTI
    & 1 & 7.58 & 5.58 & 8.45 & 7.28
    & 7.23 & 5.52 & 8.34 & 7.07 \\

    \rowcolor{gray!15}
    w/o DSA
    & 1 & 8.82 & 6.58 & 10.12 & 8.57
    & 8.88 & 7.02 & 9.76 & 8.62 \\

    \rowcolor{gray!15}
    DINOv3 $\rightarrow$ ViT
    & 1 & 6.58 & 5.32 & 8.11 & 6.63
    & 6.84 & 5.12 & 7.94 & 6.67 \\

    \bottomrule
  \end{tabularx}
  \vspace{-5mm}
\end{table}
Additional qualitative comparisons and complete 13-category quantitative
results, including image-to-3D reference comparisons, are provided in
Appendices~\ref{sec:supp_qualitative} and~\ref{sec:supp_additional_quant}, respectively.

To evaluate the practical efficiency of low-NFE generation, we compare inference time
and peak memory usage.
GPU inference is measured on a single NVIDIA RTX A4000.
After warm-up, we run each method 100 times on the same input and report the mean
and standard deviation.
We compare against PC$^2$~\cite{pc2}, BDM~\cite{bdm}, and RGB2point~\cite{rgb2point},
with the output resolution fixed to 2048 points.
Furthermore, to examine the feasibility of lightweight deployment, we additionally
evaluate Core ML inference on an iPhone 15 Pro as a supplementary experiment.
\subsubsection{Implementation Details.}
\label{sec:impl_details}
All evaluated Point-MF models are trained in PyTorch on eight NVIDIA
RTX A6000 GPUs ($\sim$25 hours); a separate JAX/XLA implementation
takes $\sim$16 hours.
Each training sample consists of an $N=2048$ point cloud and a $224\times224$ RGB image.
We use pretraining only for the image feature extractor, namely DINOv3, and train all other modules from scratch.
The DiT backbone has hidden dimension $H=512$, $L=12$ blocks, and $h=8$ attention heads.
For image conditioning, we use DINOv3 ViT-B as the frozen image encoder.
The CTI module for processing DINO patch tokens is applied once on the condition branch, with internal dimension 1024 and 4 attention heads.

The DSA weight is set to $\lambda_{\mathrm{base}}=0.5$.
We optimize the model using AdamW with learning rate $1.0\times10^{-4}$, batch size 128, and 120{,}000 training steps.
The first 10{,}000 steps are used for warm-up.
The CFG-based velocity composition used during training is included in the objective in Eq.~\eqref{eq:cfg_loss}.
Additional hyperparameter details are provided in Appendix~\ref{sec:supp_impl}.

\subsubsection{Quantitative Results.}
\label{sec:quantitative}
\begin{wraptable}[11]{r}{0.5\linewidth}
\vspace{-1.0cm}
\centering
\caption{
Inference time and peak memory usage.
GPU results are measured on an RTX A4000, and mobile results are measured on an
iPhone 15 Pro using Core ML.
}
\vskip -0.5mm
\label{tab:runtime_vram}
\scriptsize
\setlength{\tabcolsep}{2.4pt}
\renewcommand{\arraystretch}{0.95}
\resizebox{\linewidth}{!}{%
\begin{tabular}{lccc}
\toprule
Method & NFE & Time [ms/sample] & Peak Mem. \\
\midrule
\multicolumn{4}{l}{\textit{Device: RTX A4000}} \\
PC$^2$~\cite{pc2} & 256 & 22000 $\pm$ 200 & 1.730 GiB \\
BDM~\cite{bdm} & 276 & 28000 $\pm$ 200 & 1.970 GiB \\
RGB2point~\cite{rgb2point} & 1 & 28.39 $\pm$ 2.18 & 0.818 GiB \\
Ours (2048 points) & 1 & 40.72 $\pm$ 2.74 & 1.033 GiB \\
Ours (8192 points) & 1 & 84.61 $\pm$ 1.21 & 1.097 GiB \\
\midrule
\multicolumn{4}{l}{\textit{Device: iPhone 15 Pro (Core ML)}} \\
Ours (2048 points) & 1 & 281.74 $\pm$ 1.51 & 0.608 GiB \\
\bottomrule
\end{tabular}
}
\vspace{-5mm}
\end{wraptable}
Table~\ref{tab:compare_cd_emd} reports CD and EMD on ShapeNet-R2N2,
and Table~\ref{tab:compare_fs} compares F-Score.
Results on Pix3D are shown in Table~\ref{tab:compare_cd_emd_pix3d}.
As shown in Table~\ref{tab:compare_cd_emd}, Point-MF outperforms diffusion-based
methods that require multi-step sampling, while using only one NFE.

RGB2point is trained with a CD objective, and therefore has an advantage on CD
because its training objective is aligned with the evaluation metric.

However, compared with probabilistic generative methods such as PC$^2$, BDM, and
Point-MF, RGB2point performs worse in terms of EMD.
As shown in Table~\ref{tab:compare_fs}, Point-MF achieves the best average F-Score
among the compared methods. Notably, Point-MF retains strong performance when directly generating 8192 points
without additional training or RepKPU upsampling.
On the real-image Pix3D dataset, Point-MF also achieves the best overall performance
with only one NFE, as shown in Table~\ref{tab:compare_cd_emd_pix3d}.
Table~\ref{tab:runtime_vram} compares inference time and peak memory usage.
Point-MF is 1.43$\times$ slower than RGB2point (40.72 vs. 28.39 ms), but over
500$\times$ faster than PC$^2$ and BDM under the same RTX A4000 protocol.
Runtime increases when directly generating 8192 points due to the larger point-token count.
We also convert Point-MF to Core ML and verify its execution on an iPhone 15 Pro.
This mobile evaluation is not intended as a same-environment comparison with existing
methods, but rather as a supplementary experiment showing that the 1-NFE inference
of Point-MF is feasible on a real device.
Details of the Core ML conversion and mobile measurement protocol are provided in
Appendix~\ref{sec:supp_coreml}.

\subsubsection{Qualitative Results.}
\label{sec:qualitative}

\begin{figure}[tb]
  \vskip -0mm
  \centering
  \includegraphics[width=\linewidth]{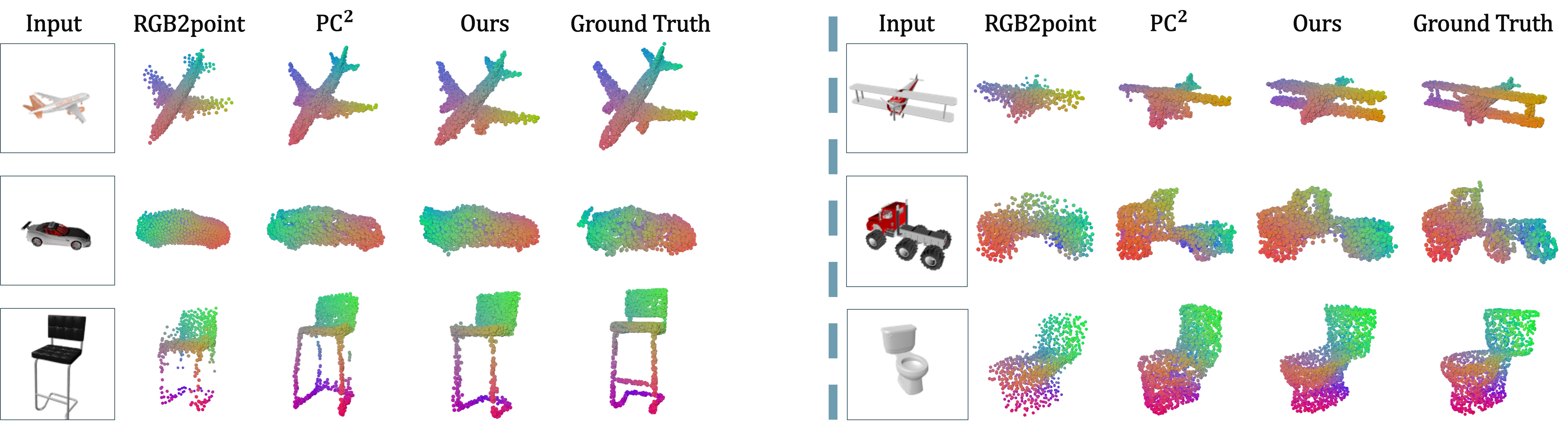}
  \caption{Qualitative results on ShapeNet.}
  \label{fig:teisei}
  \vspace{-6mm}
\end{figure}
\begin{figure}[b]
  \vspace{-5mm}
  \centering
  \includegraphics[width=1.0\linewidth]{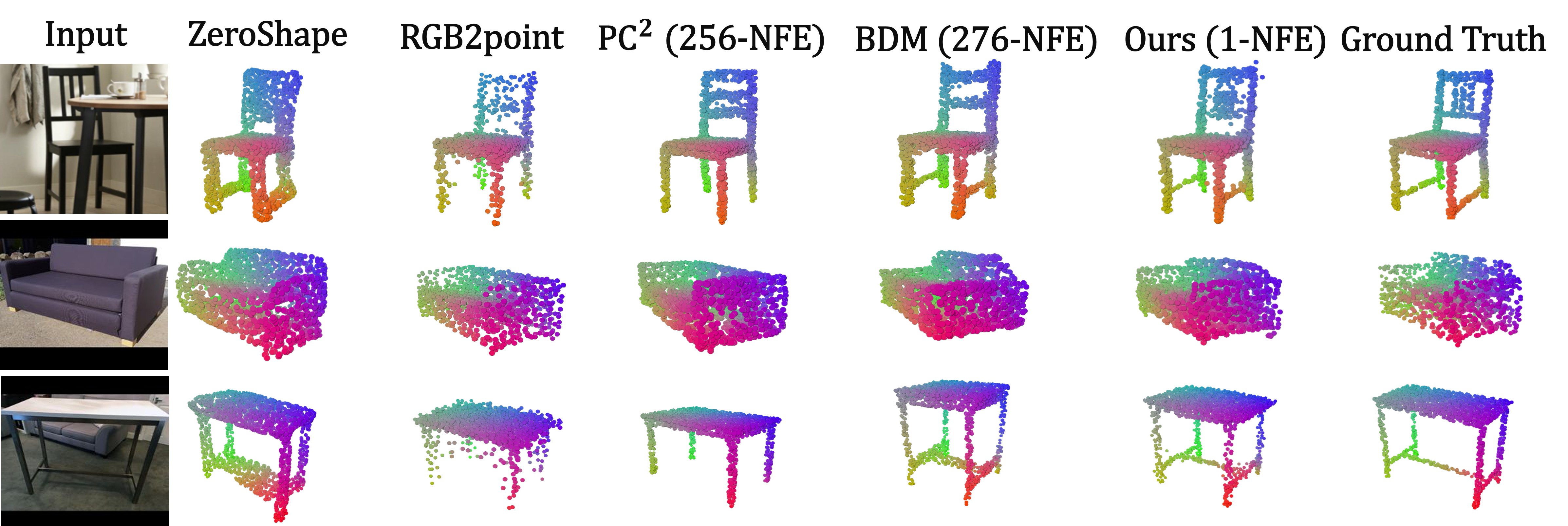}
  \caption{Qualitative results on Pix3D.}
  \label{fig:pix3d_teisei}
  \vspace{-5mm}
\end{figure}
\begin{figure}[h]
\vspace{-5mm}
  \centering
  \includegraphics[width=0.95\linewidth]{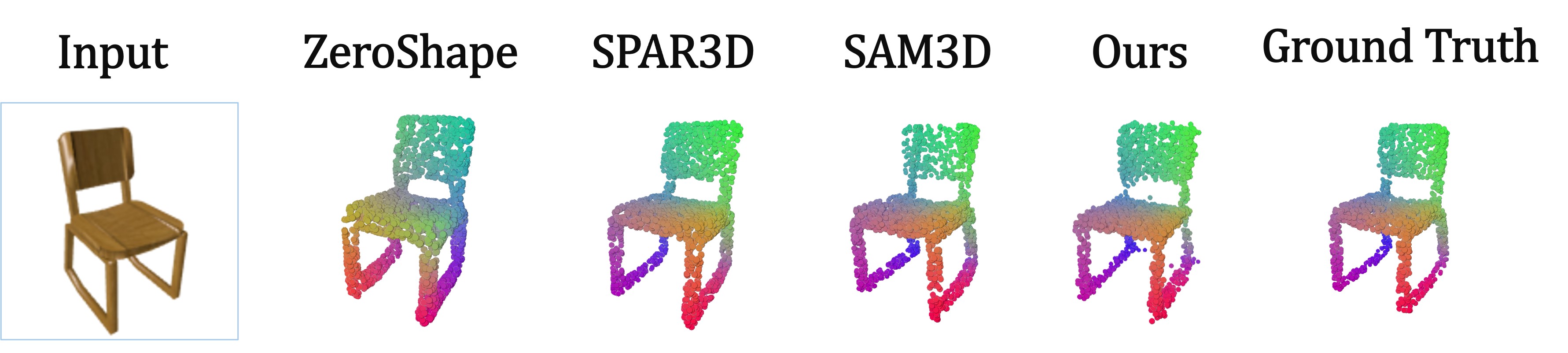}
  \caption{
Representative reference comparison with recent image-to-3D models on ShapeNet-R2N2.
All outputs are visualized as point clouds from the same viewpoint.
  }
  \label{fig:i23d_qual}
  \vspace{-5mm}
\end{figure}

To visually evaluate shape fidelity and point-density distribution,
we conduct qualitative comparisons on ShapeNet-R2N2 (Fig.~\ref{fig:teisei})
and Pix3D (Fig.~\ref{fig:pix3d_teisei}).
Overall, Point-MF generates point distributions with consistent shapes while suppressing density imbalance.
This trend is consistent with the strong EMD performance of Point-MF in Table~\ref{tab:compare_cd_emd}.
In contrast, RGB2point, which directly minimizes Chamfer Distance, often produces sharp contours because points are strongly attracted to the surface,
but it can suffer from local density imbalance or point concentration in specific regions.
On Pix3D, Point-MF can also reconstruct challenging structures such as table undersides and complex chair legs,
which are often missed by PC$^2$ and RGB2point.
We further show a subset of qualitative comparisons with recent image-to-3D models in Fig.~\ref{fig:i23d_qual}.
The qualitative results indicate that, although Point-MF does not aim to generate textures or dense meshes,
it can efficiently recover aligned object-level geometry from a single image.
Additional qualitative comparisons on the remaining ShapeNet-R2N2 categories and Pix3D,
random samples, failure cases, and additional reference comparisons with recent image-to-3D models are provided in Appendix~\ref{sec:supp_qualitative}.

\subsection{Ablations}
\label{sec:ablations}
\begin{figure}[h]
\vspace{-5mm}
  \centering
  \includegraphics[width=0.8\linewidth]{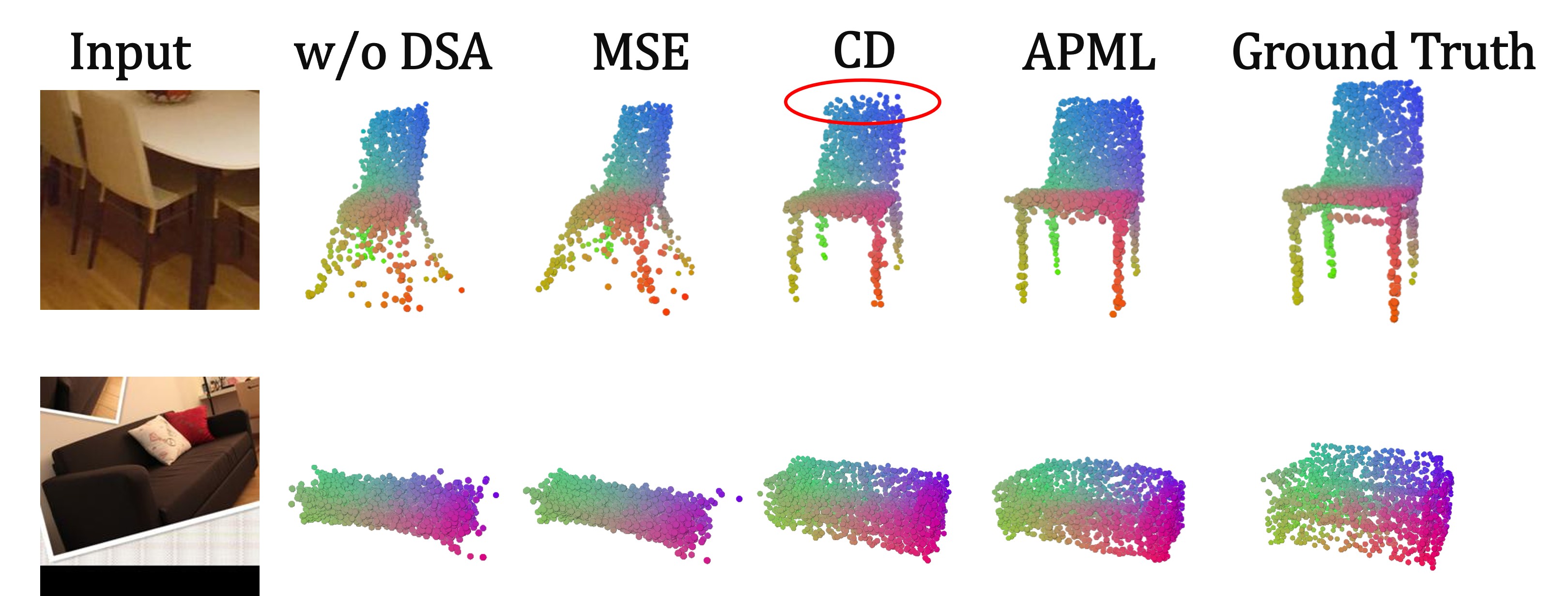}
  \caption{
  Qualitative comparison of different set distances used in DSA on Pix3D.
  When Chamfer Distance is used, points tend to scatter slightly in some regions, as indicated by the red circles.
  }
  \label{fig:ablation_setdis}
  \vspace{-5mm}
\end{figure}
Tables~\ref{tab:compare_cd_emd} and~\ref{tab:compare_cd_emd_pix3d} show that
DSA is the most influential component, CTI provides consistent gains, and replacing
DINOv3 with a standard ImageNet-pretrained ViT-B~\cite{VisionTransformer} causes only a small drop.
Thus, the gain mainly comes from denoised-space anchoring and task-specific
conditioning, not the image encoder alone.
NFE--quality and diversity analyses support our 1-NFE setting and show
retained output stochasticity under DSA (Appendix~\ref{sec:supp_ablation}).
\begin{table}[b]
\vspace{0mm}
  \centering
  \caption{Ablation study of the set distance used in DSA on Pix3D.}
  \vskip -3mm
  \label{tab:compare_cd_emd_pix3d_ab_setdis}
  \scriptsize
  \setlength{\tabcolsep}{3pt}
  \begin{tabular*}{\linewidth}{@{\extracolsep{\fill}}lcccccccc}
    \toprule
     & \multicolumn{4}{c}{CD$\times 100$ $\downarrow$} & \multicolumn{4}{c}{EMD$\times 100$ $\downarrow$} \\
    \cmidrule(lr){2-5} \cmidrule(lr){6-9}
    Method & Chair & Sofa & Table & Mean & Chair & Sofa & Table & Mean \\
    \midrule
    $\mathrm{APML}$   & \underline{6.48} & \textbf{5.02} & \textbf{8.20} & \textbf{6.53} & \textbf{6.74} & \textbf{5.06} & \textbf{7.94} & \textbf{6.61} \\
    $\mathrm{CD}$     & \textbf{6.36}    & \underline{5.12} & \underline{8.40} & \underline{6.54} & \underline{7.48} & \underline{5.26} & \underline{8.53} & \underline{7.17} \\
    $\mathrm{MSE}$    & 11.18            & 8.63            & 12.84            & 10.93            & 10.1             & 9.52             & 10.21            & 9.98 \\
    w/o DSA           & 8.82             & 6.58            & 10.12            & 8.57             & 8.88             & 7.02             & 9.76             & 8.62 \\
    \bottomrule
  \end{tabular*}
  \vspace{-5mm}
\end{table}
We further ablate the set distance used in DSA on Pix3D by replacing APML with
L1-Chamfer Distance (CD) or MSE. As shown in
Table~\ref{tab:compare_cd_emd_pix3d_ab_setdis}, APML achieves the best overall
performance, whereas MSE performs the worst because DSA regularizes an
unordered denoised point set without explicit point-to-point correspondences.
Thus, set-to-set losses are more appropriate than point-wise MSE, and APML gives
the best EMD-oriented trade-off. Fig.~\ref{fig:ablation_setdis} further shows
that CD can produce scattered points due to nearest-neighbor matching.

\subsection{Limitations}
Our method has three limitations.
First, the self-consistent target computation in Mean Flow requires a JVP,
which increases the computational cost and memory usage during training.
Second, since our method operates directly in point-cloud space, the computational
cost increases with the number of output points; obtaining higher-fidelity dense
outputs can further benefit from external upsampling or refinement methods such
as RepKPU~\cite{kpu}.
Third, our evaluation is limited to ShapeNet-R2N2 and Pix3D,
and validation on broader categories and more complex real-world images remains future work.

\section{Conclusion and Discussion}
In this paper, we formulated geometry-only point cloud reconstruction from a single image
as one-step conditional Mean Flow directly in point-cloud space,
and proposed Point-MF.
Point-MF predicts an interval-averaged velocity field without using a point-cloud autoencoder
and generates a point cloud with a single network function evaluation.
We further introduced Denoised Space Anchor (DSA),
which anchors the denoised estimate induced by the predicted mean velocity to the ground-truth shape using a set-distance loss,
thereby suppressing outliers and density imbalance caused by large one-step updates.

Experiments on ShapeNet-R2N2 and Pix3D showed that Point-MF achieves strong CD, EMD, and F-Score
against the evaluated point-cloud reconstruction baselines.
In particular, Point-MF operates with only one NFE,
while being substantially faster than iterative diffusion-based methods
and maintaining the same order of inference latency as feedforward methods.
These results demonstrate that geometric anchoring in the denoised space is effective for stabilizing one-step Mean Flow in point-cloud space.

Future work includes reducing the training cost associated with JVP,
extending Point-MF to high-resolution point cloud generation,
and validating it on broader categories and more complex real-world images.
In addition, incorporating recent improvements from Improved Mean Flows~\cite{imf}
and Pixel Mean Flow~\cite{pmf} may further improve guidance and training objectives.
\subsubsection*{Acknowledgements}
This work was supported by JST CRONOS JPMJCS24K4 and JSPS KAKENHI 26K23875.

% ---- Bibliography ----
%
% BibTeX users should specify bibliography style 'splncs04'.
% References will then be sorted and formatted in the correct style.
%

\clearpage
\FloatBarrier
\bibliographystyle{splncs04}
\bibliography{main}

@String(CVPR  = {IEEE Conf. Comput. Vis. Pattern Recog.})

@String(ICCV  = {Int. Conf. Comput. Vis.})

@String(ECCV  = {Eur. Conf. Comput. Vis.})

@String(NeurIPS = {Adv. Neural Inform. Process. Syst.})

@String(ICML  = {Int. Conf. Mach. Learn.})

@String(ICLR  = {Int. Conf. Learn. Represent.})

@String(CVPRW = {IEEE Conf. Comput. Vis. Pattern Recog. Worksh.})

@String(IJCAI = {IJCAI})

@String(TOG   = {ACM Trans. Graph.})

@String(CVPR  = {CVPR})

@String(ICCV  = {ICCV})

@String(ECCV  = {ECCV})

@String(NeurIPS = {NeurIPS})

@String(ICML  = {ICML})

@String(ICLR  = {ICLR})

@String(CVPRW = {CVPRW})

@String(TOG   = {ACM TOG})

@String(TOG    = {ACM Transactions on Graphics})

@String(CVPR   = {Proc. of the IEEE/CVF Conference on Computer Vision and Pattern Recognition})

@String(ICCV   = {Proc. of the IEEE/CVF International Conference on Computer Vision})

@String(ECCV   = {Proc. of the European Conference on Computer Vision})

@String(NeurIPS= {Advances in Neural Information Processing Systems})

@String(ICML   = {Proc. of the International Conference on Machine Learning})

@String(ICLR   = {Proc. of the International Conference on Learning Representations})

@String(CVPRW  = {Proc. of the IEEE/CVF Conference on Computer Vision and Pattern Recognition Workshops})

@String(WACV   = {Proc. of the IEEE/CVF Winter Conference on Applications of Computer Vision})

@String(IJCAI  = {Proc. of the International Joint Conference on Artificial Intelligence})

@String(THRDV    = {Proc. of the International Conference on 3D Vision})

@article{VisionTransformer,
  title={An image is worth 16x16 words: Transformers for image recognition at scale},
  author={Dosovitskiy, Alexey},
  journal={arXiv:2010.11929},
  year={2020}
}

@article{ddpm,
  title={Denoising diffusion probabilistic models},
  author={Ho, Jonathan and Jain, Ajay and Abbeel, Pieter},
  journal=neurips,
  volume={33},
  pages={6840--6851},
  year={2020}
}

@article{pointe,
  title={{Point-e}: A system for generating 3d point clouds from complex prompts},
  author={Nichol, Alex and Jun, Heewoo and Dhariwal, Prafulla and Mishkin, Pamela and Chen, Mark},
  journal={arXiv:2212.08751},
  year={2022}
}

@inproceedings{rgb2point,
  title={{Rgb2point}: {3d} point cloud generation from single rgb images},
  author={Lee, Jae Joong and Benes, Bedrich},
  booktitle=wacv,
  pages={2952--2962},
  year={2025}
}

@inproceedings{flowmatching,
  title={Flow Matching for Generative Modeling},
  author={Lipman, Yaron and Chen, Ricky TQ and Ben-Hamu, Heli and Nickel, Maximilian and Le, Matt},
  booktitle=iclr,
  year={2023}
}

@inproceedings{meanflow,
  title={Mean flows for one-step generative modeling},
  author={Geng, Zhengyang and Deng, Mingyang and Bai, Xingjian and Kolter, J Zico and He, Kaiming},
  booktitle=neurips,
  year={2025},
  volume={38},
}

@article{shapenet,
  title       = {{ShapeNet: An Information-Rich 3D Model Repository}},
  author      = {Chang, Angel X. and Funkhouser, Thomas and Guibas, Leonidas and Hanrahan, Pat and Huang, Qixing and Li, Zimo and Savarese, Silvio and Savva, Manolis and Song, Shuran and Su, Hao and Xiao, Jianxiong and Yi, Li and Yu, Fisher},
  journal      = {arXiv:1512.03012},
  year        = {2015}
}

@inproceedings{pix3d,
  title={{Pix3D}: Dataset and Methods for Single-Image 3D Shape Modeling},
  author={Sun, Xingyuan and Wu, Jiajun and Zhang, Xiuming and Zhang, Zhoutong and Zhang, Chengkai and Xue, Tianfan and Tenenbaum, Joshua B and Freeman, William T},
  booktitle=cvpr,
  pages={2974--2983},
  year={2018},
}

@inproceedings{DiT,
  title={Scalable diffusion models with transformers},
  author={Peebles, William and Xie, Saining},
  booktitle=iccv,
  pages={4195--4205},
  year={2023}
}

@article{DINO3,
  title={Dinov3},
  author={Sim{\'e}oni, Oriane and Vo, Huy V and Seitzer, Maximilian and Baldassarre, Federico and Oquab, Maxime and Jose, Cijo and Khalidov, Vasil and Szafraniec, Marc and Yi, Seungeun and Ramamonjisoa, Micha{\"e}l and others},
  journal={arXiv:2508.10104},
  year={2025}
}

@inproceedings{cfg,
  title={Classifier-Free Diffusion Guidance},
  author={Ho, Jonathan and Salimans, Tim},
  booktitle={NeurIPS 2021 Workshop on Deep Generative Models and Downstream Applications},
  year={2021}
}

@inproceedings{APML,
  title={{APML:} Adaptive Probabilistic Matching Loss for Robust 3D Point Cloud Reconstruction},
  author={Sharifipour, Sasan and Casado, Constantino {\'A}lvarez and Sabokrou, Mohammad and L{\'o}pez, Miguel Bordallo},
  booktitle=neurips,
  year={2025},
  volume={38},
}

@inproceedings{3dr2n2,
  title={3d-r2n2: A unified approach for single and multi-view 3d object reconstruction},
  author={Choy, Christopher B and Xu, Danfei and Gwak, JunYoung and Chen, Kevin and Savarese, Silvio},
  booktitle=eccv,
  pages={628--644},
  year={2016},
}

@inproceedings{selfsup,
  title={From image collections to point clouds with self-supervised shape and pose networks},
  author={Navaneet, KL and Mathew, Ansu and Kashyap, Shashank and Hung, Wei-Chih and Jampani, Varun and Babu, R Venkatesh},
  booktitle=cvpr,
  pages={1132--1140},
  year={2020}
}

@InProceedings{differ,
    author = {L Navaneet, K and Mandikal, Priyanka and Jampani, Varun and Babu, Venkatesh},
    title = {DIFFER: Moving Beyond 3D Reconstruction with Differentiable Feature Rendering},
    booktitle = cvprw,
    year = {2019}
}

@inproceedings{ulsp,
  title = {Unsupervised Learning of Shape and Pose with Differentiable Point Clouds},
  booktitle = neurips,
  author = {Insafutdinov, Eldar and Dosovitskiy, Alexey},
  year = {2018},
  volume = {31},
}

@inproceedings{lion,
  title={{LION}: latent point diffusion models for 3D shape generation},
  author={Zeng, Xiaohui and Vahdat, Arash and Williams, Francis and Gojcic, Zan and Litany, Or and Fidler, Sanja and Kreis, Karsten},
  booktitle=neurips,
  pages={10021--10039},
  year={2022}
}

@inproceedings{ga,
  title={{GaussianAnything}: Interactive Point Cloud Latent Diffusion for {3D} Generation},
  author={Lan, Yushi and Zhou, Shangchen and Lyu, Zhaoyang and Hong, Fangzhou and Yang, Shuai and Dai, Bo and Pan, Xingang and Loy, Chen Change},
  year={2025},
  booktitle=iclr,
}

@article{lego,
  title={{Legoformer}: Transformers for block-by-block multi-view 3d reconstruction},
  author={Yagubbayli, Farid and Wang, Yida and Tonioni, Alessio and Tombari, Federico},
  journal={arXiv:2106.12102},
  year={2021},
  pages={}
}

@inproceedings{pc2,
  title={{Pc2}: Projection-conditioned point cloud diffusion for single-image 3d reconstruction},
  author={Melas-Kyriazi, Luke and Rupprecht, Christian and Vedaldi, Andrea},
  booktitle=cvpr,
  pages={12923--12932},
  year={2023}
}

@inproceedings{kpu,
  title={{RepKPU}: Point Cloud Upsampling with Kernel Point Representation and Deformation},
  author={Rong, Yi and Zhou, Haoran and Xia, Kang and Mei, Cheng and Wang, Jiahao and Lu, Tong},
  booktitle=cvpr,
  pages={21050--21060},
  year={2024}
}

@article{3dgs,
  title={3d gaussian splatting for real-time radiance field rendering.},
  author={Kerbl, Bernhard and Kopanas, Georgios and Leimk{\"u}hler, Thomas and Drettakis, George and others},
  journal=TOG,
  volume={42},
  number={4},
  pages={139--1},
  year={2023}
}

@inproceedings{bdm,
  title={Bayesian diffusion models for 3d shape reconstruction},
  author={Xu, Haiyang and Lei, Yu and Chen, Zeyuan and Zhang, Xiang and Zhao, Yue and Wang, Yilin and Tu, Zhuowen},
  booktitle=cvpr,
  pages={10628--10638},
  year={2024}
}

@inproceedings{deepunsup,
  title={Deep unsupervised learning using nonequilibrium thermodynamics},
  author={Sohl-Dickstein, Jascha and Weiss, Eric and Maheswaranathan, Niru and Ganguli, Surya},
  booktitle=icml,
  pages={2256--2265},
  year={2015}
}

@inproceedings{edm,
 author = {Karras, Tero and Aittala, Miika and Aila, Timo and Laine, Samuli},
 booktitle = neurips,
 pages = {26565--26577},
 title = {Elucidating the Design Space of Diffusion-Based Generative Models},
 volume = {35},
 year = {2022}
}

@article{scoresde,
  title={Generative modeling by estimating gradients of the data distribution},
  author={Song, Yang and Ermon, Stefano},
  journal=neurips,
  volume={32},
  year={2019}
}

@inproceedings{scorematch,
  title={Score-Based Generative Modeling through Stochastic Differential Equations},
  author={Song, Yang and Sohl-Dickstein, Jascha and Kingma, Diederik P and Kumar, Abhishek and Ermon, Stefano and Poole, Ben},
  booktitle=iclr,
  year={2021}
}

@inproceedings{recflow,
  title={Flow Straight and Fast: Learning to Generate and Transfer Data with Rectified Flow},
  author={Liu, Xingchao and Gong, Chengyue and Liu, Qiang},
  booktitle=iclr,
  year={2023}
}

@inproceedings{cm,
author = {Song, Yang and Dhariwal, Prafulla and Chen, Mark and Sutskever, Ilya},
title = {Consistency models},
year = {2023},
booktitle = icml,
numpages = {42},
}

@inproceedings{icm,
  title={Improved Techniques for Training Consistency Models},
  author={Song, Yang and Dhariwal, Prafulla},
  booktitle=iclr,
  year = {2024}
}

@inproceedings{cm_easy,
  title={Consistency Models Made Easy},
  author={Geng, Zhengyang and Pokle, Ashwini and Luo, Weijian and Lin, Justin and Kolter, J Zico},
  booktitle=iclr,
  year = {2025}
}

@inproceedings{ct_consis,
  title={Simplifying, Stabilizing and Scaling Continuous-time Consistency Models},
  author={Lu, Cheng and Song, Yang},
  booktitle=iclr,
  year = {2025}
}

@inproceedings{ctm,
  title={Consistency Trajectory Models: Learning Probability Flow ODE Trajectory of Diffusion},
  author={Kim, Dongjun and Lai, Chieh-Hsin and Liao, Wei-Hsiang and Murata, Naoki and Takida, Yuhta and Uesaka, Toshimitsu and He, Yutong and Mitsufuji, Yuki and Ermon, Stefano},
  booktitle=iclr,
  year = {2024}
}

@inproceedings{shortcut,
  title={One Step Diffusion via Shortcut Models},
  author={Frans, Kevin and Hafner, Danijar and Levine, Sergey and Abbeel, Pieter},
  booktitle=iclr,
  year={2025}
}

@inproceedings{imm,
  title={Inductive Moment Matching},
  author={Zhou, Linqi and Ermon, Stefano and Song, Jiaming},
  booktitle=icml,
  year={2025}
}

@inproceedings{alpha,
  title={Alphaflow: Understanding and improving meanflow models},
  author={Zhang, Huijie and Siarohin, Aliaksandr and Menapace, Willi and Vasilkovsky, Michael and Tulyakov, Sergey and Qu, Qing and Skorokhodov, Ivan},
  booktitle=iclr,
  year={2026}
}

@inproceedings{dmf,
  title={Decoupled meanflow: Turning flow models into flow maps for accelerated sampling},
  author={Lee, Kyungmin and Yu, Sihyun and Shin, Jinwoo},
  booktitle=iclr,
  year={2026}
}

@inproceedings{genre,
  title={3d reconstruction of novel object shapes from single images},
  author={Thai, Anh and Stojanov, Stefan and Upadhya, Vijay and Rehg, James M},
  booktitle= THRDV,
  pages={85--95},
  year={2021},
}

@inproceedings{pixel2mesh,
  title={Pixel2mesh: Generating 3d mesh models from single rgb images},
  author={Wang, Nanyang and Zhang, Yinda and Li, Zhuwen and Fu, Yanwei and Liu, Wei and Jiang, Yu-Gang},
  booktitle=eccv,
  pages={52--67},
  year={2018}
}

@inproceedings{atlas,
  title={Atlas: End-to-end 3d scene reconstruction from posed images},
  author={Murez, Zak and Van As, Tarrence and Bartolozzi, James and Sinha, Ayan and Badrinarayanan, Vijay and Rabinovich, Andrew},
  booktitle=eccv,
  pages={414--431},
  year={2020},
}

@inproceedings{atlasnet,
  title={A papier-m{\^a}ch{\'e} approach to learning 3d surface generation},
  author={Groueix, Thibault and Fisher, Matthew and Kim, Vladimir G and Russell, Bryan C and Aubry, Mathieu},
  booktitle=cvpr,
  pages={216--224},
  year={2018},
  organization={IEEE}
}

@inproceedings{occ,
  title={Occupancy networks: Learning 3d reconstruction in function space},
  author={Mescheder, Lars and Oechsle, Michael and Niemeyer, Michael and Nowozin, Sebastian and Geiger, Andreas},
  booktitle=cvpr,
  pages={4460--4470},
  year={2019}
}

@inproceedings{deepsdf,
  title={Deepsdf: Learning continuous signed distance functions for shape representation},
  author={Park, Jeong Joon and Florence, Peter and Straub, Julian and Newcombe, Richard and Lovegrove, Steven},
  booktitle=cvpr,
  pages={165--174},
  year={2019}
}

@inproceedings{gsd,
  title={Gsd: View-guided gaussian splatting diffusion for 3d reconstruction},
  author={Mu, Yuxuan and Zuo, Xinxin and Guo, Chuan and Wang, Yilin and Lu, Juwei and Wu, Xiaofeng and Xu, Songcen and Dai, Peng and Yan, Youliang and Cheng, Li},
  booktitle=eccv,
  pages={55--72},
  year={2024},
}

@article{dit3d,
  title={Dit-3d: Exploring plain diffusion transformers for 3d shape generation},
  author={Mo, Shentong and Xie, Enze and Chu, Ruihang and Hong, Lanqing and Niessner, Matthias and Li, Zhenguo},
  journal=neurips,
  volume={36},
  pages={67960--67971},
  year={2023}
}

@inproceedings{pc2gen,
  title={Diffusion probabilistic models for 3d point cloud generation},
  author={Luo, Shitong and Hu, Wei},
  booktitle=cvpr,
  pages={2837--2845},
  year={2021}
}

@inproceedings{spar3d,
  title={Spar3d: Stable point-aware reconstruction of 3d objects from single images},
  author={Huang, Zixuan and Boss, Mark and Vasishta, Aaryaman and Rehg, James M and Jampani, Varun},
  booktitle=cvpr,
  pages={16860--16870},
  year={2025}
}

@inproceedings{pointnerf,
  title={Point-nerf: Point-based neural radiance fields},
  author={Xu, Qiangeng and Xu, Zexiang and Philip, Julien and Bi, Sai and Shu, Zhixin and Sunkavalli, Kalyan and Neumann, Ulrich},
  booktitle=cvpr,
  pages={5438--5448},
  year={2022}
}

@inproceedings{poco,
  title={Poco: Point convolution for surface reconstruction},
  author={Boulch, Alexandre and Marlet, Renaud},
  booktitle=cvpr,
  pages={6302--6314},
  year={2022}
}

@InProceedings{sdfcon,
    author    = {Jignasu, Anushrut and Balu, Aditya and Sarkar, Soumik and Hegde, Chinmay and Ganapathysubramanian, Baskar and Krishnamurthy, Adarsh},
    title     = {{SDFConnect}: Neural Implicit Surface Reconstruction of a Sparse Point Cloud with Topological Constraints},
    booktitle = cvprw,
    year      = {2024},
    pages     = {5271-5279}
}

@inproceedings{sf3d,
  title={Sf3d: Stable fast 3d mesh reconstruction with uv-unwrapping and illumination disentanglement},
  author={Boss, Mark and Huang, Zixuan and Vasishta, Aaryaman and Jampani, Varun},
  booktitle=cvpr,
  pages={16240--16250},
  year={2025}
}

@inproceedings{lrm,
  title={{LRM}: Large Reconstruction Model for Single Image to 3D},
  author={Hong, Yicong and Zhang, Kai and Gu, Jiuxiang and Bi, Sai and Zhou, Yang and Liu, Difan and Liu, Feng and Sunkavalli, Kalyan and Bui, Trung and Tan, Hao},
  booktitle=iclr,
  year={2024}
}

@INPROCEEDINGS{shapeclip,
  author={Huang, Zixuan and Jampani, Varun and Thai, Anh and Li, Yuanzhen and Stojanov, Stefan and Rehg, James M.},
  booktitle=cvpr, 
  title={{ShapeClipper}: Scalable {3D} Shape Learning from Single-View Images via Geometric and CLIP-Based Consistency}, 
  year={2023},
  pages={12912-12922},
  }

@inproceedings{crm,
  title={Crm: Single image to 3d textured mesh with convolutional reconstruction model},
  author={Wang, Zhengyi and Wang, Yikai and Chen, Yifei and Xiang, Chendong and Chen, Shuo and Yu, Dajiang and Li, Chongxuan and Su, Hang and Zhu, Jun},
  booktitle=eccv,
  pages={57--74},
  year={2024}
}

@inproceedings{chen2023single,
  title={Single-stage diffusion nerf: A unified approach to 3d generation and reconstruction},
  author={Chen, Hansheng and Gu, Jiatao and Chen, Anpei and Tian, Wei and Tu, Zhuowen and Liu, Lingjie and Su, Hao},
  booktitle=cvpr,
  pages={2416--2425},
  year={2023}
}

@inproceedings{diffusionsddf,
  title={Diffusion-sdf: Conditional generative modeling of signed distance functions},
  author={Chou, Gene and Bahat, Yuval and Heide, Felix},
  booktitle=iccv,
  pages={2262--2272},
  year={2023}
}

@inproceedings{pointinfinity,
  title={Pointinfinity: Resolution-invariant point diffusion models},
  author={Huang, Zixuan and Johnson, Justin and Debnath, Shoubhik and Rehg, James M and Wu, Chao-Yuan},
  booktitle=cvpr,
  pages={10050--10060},
  year={2024}
}

@article{shap3,
  title={Shap-e: Generating conditional 3d implicit functions},
  author={Jun, Heewoo and Nichol, Alex},
  journal={arXiv preprint arXiv:2305.02463},
  year={2023}
}

@inproceedings{ln3diff,
  title={{ln3Diff}: Scalable Latent Neural Fields Diffusion for Speedy 3D Generation},
  author={Lan, Yushi and Hong, Fangzhou and Yang, Shuai and Zhou, Shangchen and Meng, Xuyi and Dai, Bo and Pan, Xingang and Loy, Chen Change},
  booktitle=eccv,
  pages={112--130},
  year={2024},
  organization={Springer}
}

@article{imf,
  title={Improved mean flows: On the challenges of fastforward generative models},
  author={Geng, Zhengyang and Lu, Yiyang and Wu, Zongze and Shechtman, Eli and Kolter, J Zico and He, Kaiming},
  journal={arXiv preprint arXiv:2512.02012},
  year={2025}
}

@inproceedings{pmf,
  title={One-step Latent-free Image Generation with Pixel Mean Flows},
  author={Lu, Yiyang and Lu, Susie and Sun, Qiao and Zhao, Hanhong and Jiang, Zhicheng and Wang, Xianbang and Li, Tianhong and Geng, Zhengyang and He, Kaiming},
  booktitle={arXiv preprint arXiv:2601.22158},
  year={2026}
}

@inproceedings{bring,
  title={Bring metric functions into diffusion models},
  author={An, Jie and Yang, Zhengyuan and Wang, Jianfeng and Li, Linjie and Liu, Zicheng and Wang, Lijuan and Luo, Jiebo},
  booktitle=ijcai,
  pages={578--586},
  year={2024}
}

@article{irf,
  title={Improving the training of rectified flows},
  author={Lee, Sangyun and Lin, Zinan and Fanti, Giulia},
  journal=neurips,
  volume={37},
  pages={63082--63109},
  year={2024}
}

@inproceedings{trellis,
  title={Structured 3d latents for scalable and versatile 3d generation},
  author={Xiang, Jianfeng and Lv, Zelong and Xu, Sicheng and Deng, Yu and Wang, Ruicheng and Zhang, Bowen and Chen, Dong and Tong, Xin and Yang, Jiaolong},
  booktitle=cvpr,
  pages={21469--21480},
  year={2025}
}

@inproceedings{trellis2,
  title={Native and compact structured latents for 3d generation},
  author={Xiang, Jianfeng and Chen, Xiaoxue and Xu, Sicheng and Wang, Ruicheng and Lv, Zelong and Deng, Yu and Zhu, Hongyuan and Dong, Yue and Zhao, Hao and Yuan, Nicholas Jing and others},
  booktitle=cvpr,
  pages={14419--14429},
  year={2026}
}

@inproceedings{sparseflex,
  title={{Sparseflex}: High-resolution and arbitrary-topology 3d shape modeling},
  author={He, Xianglong and Zou, Zi-Xin and Chen, Chia-Hao and Guo, Yuan-Chen and Liang, Ding and Yuan, Chun and Ouyang, Wanli and Cao, Yan-Pei and Li, Yangguang},
  booktitle=cvpr,
  pages={14822--14833},
  year={2025}
}

@inproceedings{lattice,
  title={Lattice: Democratize high-fidelity 3d generation at scale},
  author={Lai, Zeqiang and Zhao, Yunfei and Zhao, Zibo and Liu, Haolin and Lin, Qingxiang and Huang, Jingwei and Guo, Chunchao and Yue, Xiangyu},
  booktitle=cvpr,
  pages={19982--19992},
  year={2026}
}

@inproceedings{ultrashape,
  title={{UltraShape 1.0}: High-Fidelity {3D} Shape Generation via Scalable Geometric Refinement},
  author={Jia, Tanghui and Yan, Dongyu and Hao, Dehao and Li, Yang and Zhang, Kaiyi and He, Xianyi and Li, Lanjiong and Wang, Yuhan and Chen, Jinnan and Jiang, Lutao and others},
  booktitle={arXiv preprint arXiv:2512.21185},
  year={2025}
}

@inproceedings{sam3d,
  title={Sam 3d: 3dfy anything in images},
  author={Chen, Xingyu and Chu, Fu-Jen and Gleize, Pierre and Liang, Kevin J and Sax, Alexander and Tang, Hao and Wang, Weiyao and Guo, Michelle and Hardin, Thibaut and Li, Xiang and others},
  booktitle=cvpr,
  pages={7220--7232},
  year={2026}
}

@inproceedings{zeroshape,
  title={Zeroshape: Regression-based zero-shot shape reconstruction},
  author={Huang, Zixuan and Stojanov, Stefan and Thai, Anh and Jampani, Varun and Rehg, James M},
  booktitle=cvpr,
  pages={10061--10071},
  year={2024}
}

@inproceedings{zerow,
  title={Robust 3d shape reconstruction in zero-shot from a single image in the wild},
  author={Cho, Junhyeong and Youwang, Kim and Yang, Hunmin and Oh, Tae-Hyun},
  booktitle=cvpr,
  pages={22786--22798},
  year={2025}
}

@article{fcm,
  title={Adaptive 3D Reconstruction via Diffusion Priors and Forward Curvature-Matching Likelihood Updates},
  author={Shin, Seunghyeok and Kim, Dabin and Lim, Hongki},
  journal=neurips,
  volume={38},
  pages={128819--128856},
  year={2025}
}

@inproceedings{f2d2,
  title={Joint distillation for fast likelihood evaluation and sampling in flow-based models},
  author={Ai, Xinyue and He, Yutong Kelly and Gu, Albert and Salakhutdinov, Russ and Kolter, Zico and Boffi, Nicholas and Simchowitz, Max},
  booktitle=iclr,
  volume={2026},
  pages={44940--44977},
  year={2026}
}

@inproceedings{graspnet,
  author = {Fang, Hao-Shu and Wang, Chenxi and Gou, Minghao and Lu, Cewu},
  title = {{GraspNet-1Billion}: A Large-Scale Benchmark for General Object Grasping},
  booktitle = cvpr,
  pages = {11444--11453},
  year = {2020}
}

% ============================================================
% SUPPLEMENTARY MATERIAL
% ============================================================
\clearpage
\appendix

% Restore the standalone supplement's subsection styles. The paper title
% and running heads remain those of the main paper; no second \maketitle.
\let\subsection\PointMFSuppSubsection
\let\subsubsection\PointMFSuppSubsubsection

% Reset supplementary counters once, not at every appendix section.
% S prefixes distinguish these objects from the main-paper objects.
% Do NOT reset page numbers or bibliography numbering.
\setcounter{figure}{0}
\setcounter{table}{0}
\setcounter{equation}{0}
\setcounter{algoboxctr}{0}
\renewcommand{\thefigure}{S\arabic{figure}}
\renewcommand{\thetable}{S\arabic{table}}
\renewcommand{\theequation}{S\arabic{equation}}
\renewcommand{\thealgoboxctr}{S\arabic{algoboxctr}}

% PDF link targets must also be unique after counter resets.
% LNCS appendix section anchors are made explicitly distinct as well.
\renewcommand*{\theHsection}{supp.\Alph{section}}
\renewcommand*{\theHsubsection}{\theHsection.\arabic{subsection}}
\renewcommand*{\theHsubsubsection}{\theHsubsection.\arabic{subsubsection}}
\renewcommand*{\theHfigure}{supp.\arabic{figure}}
\renewcommand*{\theHtable}{supp.\arabic{table}}
\renewcommand*{\theHequation}{supp.\arabic{equation}}
\providecommand*{\theHalgoboxctr}{}
\renewcommand*{\theHalgoboxctr}{supp.\arabic{algoboxctr}}
\section*{Supplementary Material}

\section{CFG-Guided Mean Flow}
\label{sec:supp_cfg_mean_flow}

\subsubsection{Derivation of the Training Objective}
To strengthen generation aligned with the condition $c$, we incorporate
Classifier-Free Guidance (CFG)~\cite{cfg} into Mean Flow~\cite{meanflow}.
In this work, CFG is not applied at inference time by evaluating conditional
and unconditional branches twice. Instead, we incorporate it into the Mean Flow
training target. Therefore, it does not increase the NFE during sampling.

Let $v(x_t,t\mid c)$ denote the conditional instantaneous velocity field, and
let $v(x_t,t)$ denote the unconditional instantaneous velocity field.
The CFG-guided instantaneous velocity field is defined as
\begin{align}
v_{\mathrm{cfg}}(x_t,t\mid c)
\coloneqq
\omega\,v(x_t,t\mid c)
+
(1-\omega)\,v(x_t,t).
\label{eq:supp_cfg_inst_field}
\end{align}
Here, $\omega$ is the CFG coefficient.
In the standard case without the $\kappa$ term introduced later,
$\omega$ directly corresponds to the CFG scale.

Averaging this velocity field over the interval $0\le r<t\le 1$ gives the
CFG-guided mean velocity field:
\begin{align}
u_{\mathrm{cfg}}(x_t,r,t\mid c)
\coloneqq
\frac{1}{t-r}
\int_{r}^{t}
v_{\mathrm{cfg}}(x_\tau,\tau\mid c)\,d\tau .
\label{eq:supp_cfg_mean_vel_def}
\end{align}
Multiplying both sides of Eq.~\eqref{eq:supp_cfg_mean_vel_def} by $(t-r)$
and differentiating with respect to $t$ while treating $r$ as a constant,
i.e., with $\frac{dr}{dt}=0$, yields
\begin{align}
u_{\mathrm{cfg}}(x_t,r,t\mid c)
=
v_{\mathrm{cfg}}(x_t,t\mid c)
-
(t-r)
\frac{d}{dt}
u_{\mathrm{cfg}}(x_t,r,t\mid c).
\label{eq:supp_cfg_identity}
\end{align}

Let the tangent vector be $\dot{x}_t \coloneqq \frac{dx_t}{dt}$.
The total derivative can then be written as
\begin{align}
\frac{d}{dt}
u_{\mathrm{cfg}}(x_t,r,t\mid c)
=
\partial_x u_{\mathrm{cfg}}(x_t,r,t\mid c)\,\dot{x}_t
+
\partial_r u_{\mathrm{cfg}}(x_t,r,t\mid c)\frac{dr}{dt}
+
\partial_t u_{\mathrm{cfg}}(x_t,r,t\mid c).
\label{eq:supp_cfg_total_derivative_full}
\end{align}
Since $r$ is treated as independent of $t$, we have $\frac{dr}{dt}=0$, and thus
\begin{align}
\frac{d}{dt}
u_{\mathrm{cfg}}(x_t,r,t\mid c)
=
\partial_x u_{\mathrm{cfg}}(x_t,r,t\mid c)\,\dot{x}_t
+
\partial_t u_{\mathrm{cfg}}(x_t,r,t\mid c).
\label{eq:supp_cfg_total_derivative}
\end{align}

Under the true generation dynamics, Eq.~\eqref{eq:supp_cfg_inst_field} gives
$\dot{x}_t=v_{\mathrm{cfg}}(x_t,t\mid c)$.
However, $v(x_t,t\mid c)$ and $v(x_t,t)$ involve marginalization and are
therefore not directly observable during training.
In this paper, the ground-truth point cloud is denoted by $x^{\mathrm{GT}}_0$,
and the noise point cloud is denoted by $\epsilon\sim\mathcal{N}(0,I)$.
We use the time convention in which $t=0$ corresponds to the data side and
$t=1$ corresponds to the noise side.
Then, the state at time $t$ and the path-wise velocity are given by
\begin{align}
x_t
=
(1-t)x^{\mathrm{GT}}_0
+
t\epsilon,
\qquad
v_t
=
\epsilon
-
x^{\mathrm{GT}}_0 .
\label{eq:supp_xt_vt}
\end{align}
We use this $v_t$ as a sample-conditional surrogate supervision signal for the
conditional instantaneous velocity field.

\paragraph{Improved CFG target.}
Mean Flow learns the CFG-guided mean velocity field $u_{\mathrm{cfg}}$
directly, rather than applying CFG at inference time.
In standard CFG-guided Mean Flow, the fact that the unconditional mean
velocity field coincides with the instantaneous velocity field at $r=t$ is used
to incorporate $v_{\mathrm{cfg}}$ into the target.
Furthermore, following the Improved CFG formulation of the original
Mean Flow paper, we mix the conditional and unconditional one-timepoint
predictions.
Specifically, we define
\begin{align}
u_c
&=
u_\theta(x_t,t,t\mid c),\\
u_\varnothing
&=
u_\theta(x_t,t,t\mid \varnothing)
\end{align}
and define the approximate CFG instantaneous velocity, namely the tangent
vector used for the JVP, as
\begin{align}
\tilde{v}_t
\coloneqq
\omega\,v_t
+
\kappa\,u_c
+
(1-\omega-\kappa)\,u_\varnothing .
\label{eq:supp_cfg_vtilde}
\end{align}
Here, $\varnothing$ denotes the condition-dropped input.
The coefficient $\kappa\ge 0$ controls how much the conditional one-timepoint
prediction $u_\theta(x_t,t,t\mid c)$ is mixed into the target side.
When $\kappa=0$, the target reduces to the standard CFG-guided Mean Flow
target.

The formulation in Eq.~\eqref{eq:supp_cfg_vtilde}, which includes $\kappa$,
has the effective CFG scale
\begin{align}
\omega_{\mathrm{eff}}
=
\frac{\omega}{1-\kappa}.
\label{eq:supp_effective_cfg_scale}
\end{align}
Thus, the actual guidance strength is determined not by $\omega$ alone, but by
$\omega_{\mathrm{eff}}$.
For example, the setting used in this work, $(\omega,\kappa)=(1.0,0.5)$,
corresponds to $\omega_{\mathrm{eff}}=2.0$.

During training, as in standard Classifier-Free Guidance, we drop the condition
with probability $p_{\mathrm{drop}}$:
\begin{align}
\bar{c}
=
\begin{cases}
\varnothing, & \text{with probability } p_{\mathrm{drop}},\\
c, & \text{otherwise}.
\end{cases}
\label{eq:supp_condition_dropout}
\end{align}
The main branch output is then given by $u_\theta(x_t,r,t\mid \bar{c})$.
This allows the same network to handle both conditional and unconditional
inputs.

By replacing the tangent vector $\dot{x}_t$ in Eq.~\eqref{eq:supp_cfg_identity}
with $\tilde{v}_t$, the regression target becomes
\begin{align}
u^{\mathrm{cfg}}_{\mathrm{tgt}}
\coloneqq
\tilde{v}_t
-
(t-r)
\frac{d}{dt}
u_\theta(x_t,r,t\mid \bar{c}).
\label{eq:supp_cfg_target}
\end{align}
This total derivative can be computed efficiently using a Jacobian--vector
product (JVP).
Specifically, for the inputs $(x_t,r,t)$ of $u_\theta$, we provide the tangent
vector $(\tilde{v}_t,0,1)$:
\begin{align}
\left(
u_\theta(x_t,r,t\mid \bar{c}),
\frac{d}{dt}
u_\theta(x_t,r,t\mid \bar{c})
\right)
=
\mathrm{jvp}
\left(
u_\theta,\,
(x_t,r,t),\,
(\tilde{v}_t,0,1)
\right).
\label{eq:supp_cfg_jvp}
\end{align}

After fixing the target with stop-gradient, the Mean Flow regression error is
\begin{align}
\Delta_{\mathrm{MF}}
=
u_\theta(x_t,r,t\mid \bar{c})
-
\mathrm{sg}
\left(
u^{\mathrm{cfg}}_{\mathrm{tgt}}
\right).
\label{eq:supp_cfg_delta_mf}
\end{align}
Therefore, the basic CFG-guided Mean Flow loss is
\begin{align}
\mathcal{L}_{\mathrm{MF\text{-}CFG}}(\theta)
=
\mathbb{E}
\left[
\left\|
\Delta_{\mathrm{MF}}
\right\|_2^2
\right].
\label{eq:supp_cfg_loss}
\end{align}

When $r=t$, we have $(t-r)=0$, and thus
Eq.~\eqref{eq:supp_cfg_target} becomes
$u^{\mathrm{cfg}}_{\mathrm{tgt}}=\tilde{v}_t$.
In this case, the objective reduces to a one-timepoint regression objective
similar to Flow Matching.
In practice, by mixing the cases $r=t$ and $r\ne t$, the model learns both
local Flow-Matching-like behavior and Mean Flow updates over large intervals.

\subsubsection{Adaptive L2 Loss}

Specifically, we define
\begin{align}
L_{\mathrm{sq}}
=
\left\|
\Delta_{\mathrm{MF}}
\right\|_2^2
\end{align}
and introduce the weight
\begin{align}
w_{\mathrm{adpt}}
=
\frac{1}{
\left(
L_{\mathrm{sq}}+\eta_{\mathrm{adpt}}
\right)^p
}.
\label{eq:supp_adaptive_weight}
\end{align}
Here, $\eta_{\mathrm{adpt}}>0$ is a constant for numerical stability, and
$p$ is a hyperparameter that controls the strength of the weighting.
The final Mean Flow loss is then given by
\begin{align}
\mathcal{L}^{\mathrm{adpt}}_{\mathrm{MF\text{-}CFG}}
=
\mathbb{E}
\left[
\mathrm{sg}(w_{\mathrm{adpt}})
L_{\mathrm{sq}}
\right].
\label{eq:supp_adaptive_mf_cfg_loss}
\end{align}
Following Mean Flows~\cite{meanflow}, we use adaptive loss weighting to
avoid training being overly dominated by large errors.
This corresponds to implementing a general
$\|\Delta_{\mathrm{MF}}\|^{2\gamma}_2$-type loss as a weighted squared error.
The setting $p=0.5$ is close to a Pseudo-Huber-style weighting, but the original
Mean Flows paper compares multiple values of $p$ and reports that $p=1$
works well. Therefore, we also use $p=1$ in this work.
To avoid confusion with the condition $c$, we denote the stabilizing constant
for the loss weight by $\eta_{\mathrm{adpt}}$.
\begin{table}[tbp]
  \centering
  \begin{minipage}{\linewidth}
    \small
    \raggedright
    \setlength{\parindent}{0pt}
    \setlength{\parskip}{0pt}

    \noindent\rule{\linewidth}{0.4pt}\par
    \vspace{1pt}

    \refstepcounter{algoboxctr}\label{alg:cfg_guided_mf}
    \noindent\textbf{Algorithm \thealgoboxctr.
    CFG-Guided Mean Flow: training guidance.}\par

    \vspace{1pt}
    \noindent\rule{\linewidth}{0.4pt}\par

    \begin{algobox}
      \ttfamily\scriptsize
      \raggedright

      \textcolor{algcomment}{%
        \# fn(z, r, t, c): function that predicts the mean velocity u%
      }\par
      \textcolor{algcomment}{\# x: training batch}\par
      \textcolor{algcomment}{\# c: condition batch}\par
      \vspace{3pt}

      r, t = sample\_r\_t()\par
      e = randn\_like(x)\par
      \vspace{3pt}

      x\_t = (1 - t) * x + t * e\par
      v\_t = e - x\par
      \vspace{3pt}

      \textcolor{algcomment}{%
        \# conditional and unconditional one-timepoint predictions%
      }\par
      u\_c = fn(x\_t, t, t, c)\par
      u\_u = fn(x\_t, t, t, null)\par
      \vspace{3pt}

      \textcolor{algcomment}{%
        \# approximate CFG instantaneous velocity%
      }\par
      v\_tgt = \(\omega\) * v\_t + \(\kappa\) * u\_c
              + (1 - \(\omega\) - \(\kappa\)) * u\_u\par
      \vspace{3pt}

      \textcolor{algcomment}{%
        \# main branch uses condition dropout%
      }\par
      c\_drop = dropout\_condition(c)\par
      \vspace{3pt}

      \textcolor{algcomment}{%
        \# compute du/dt by JVP with respect to (x\_t, r, t)%
      }\par
      u, dudt = jvp(fn, (x\_t, r, t, c\_drop),\par
      \hspace*{4em}(v\_tgt, 0, 1, 0))\par
      \vspace{3pt}

      \textcolor{algcomment}{%
        \# mean-flow regression target%
      }\par
      u\_tgt = v\_tgt - (t - r) * dudt\par
      error = u - stopgrad(u\_tgt)\par
      \vspace{3pt}

      loss = AdaptiveL2(error)\par
    \end{algobox}
  \end{minipage}
\end{table}

Algorithm~\ref{alg:cfg_guided_mf} shows the computation of the training
target for CFG-guided Mean Flow.
As in the original Mean Flow training, the JVP is computed with respect to
$(x_t,r,t)$, and the tangent vector $(\tilde{v}_t,0,1)$ is used.
In addition, rather than applying stop-gradient separately to the derivative term
obtained by the JVP, we apply stop-gradient to the entire final target
$u^{\mathrm{cfg}}_{\mathrm{tgt}}$.

\subsubsection{One-Step Sampling at Inference Time}
The CFG formulation in this section is not intended to evaluate both conditional
and unconditional branches at inference time.
Instead, it is used to correct the Mean Flow training target toward the
conditional direction.
Therefore, the unconditional branch is not additionally evaluated during
inference.

Starting from initial noise $\epsilon\sim\mathcal{N}(0,I)$, we fix
$(r,t)=(0,1)$ and compute
\begin{align}
x_0
=
\epsilon
-
u_\theta(\epsilon,0,1\mid c).
\label{eq:supp_one_step_sampling}
\end{align}
Thus, the number of network function evaluations during sampling remains
1-NFE.

\subsubsection{Relationship to DSA}
The loss
$\mathcal{L}^{\mathrm{adpt}}_{\mathrm{MF\text{-}CFG}}$
defined here is the main Mean Flow regression loss in this paper.
In the final training objective, we add Denoised Space Anchor (DSA), defined
in the main paper, as an auxiliary loss.
This regularizes not only the self-consistency of the velocity field, but also
the geometric consistency of the clean-space point cloud induced by the predicted
mean velocity.
In other words, CFG-guided Mean Flow is the main objective for learning a
condition-aligned mean velocity field, while DSA geometrically stabilizes the
denoised point cloud obtained from that mean velocity field.

\section{Additional Implementation Details}
\label{sec:supp_impl}

\subsubsection{Details of the DiT Block}
\begin{wrapfigure}[25]{r}{0.50\linewidth}
\vspace{-1.2\baselineskip}
\centering
\includegraphics[width=0.98\linewidth]{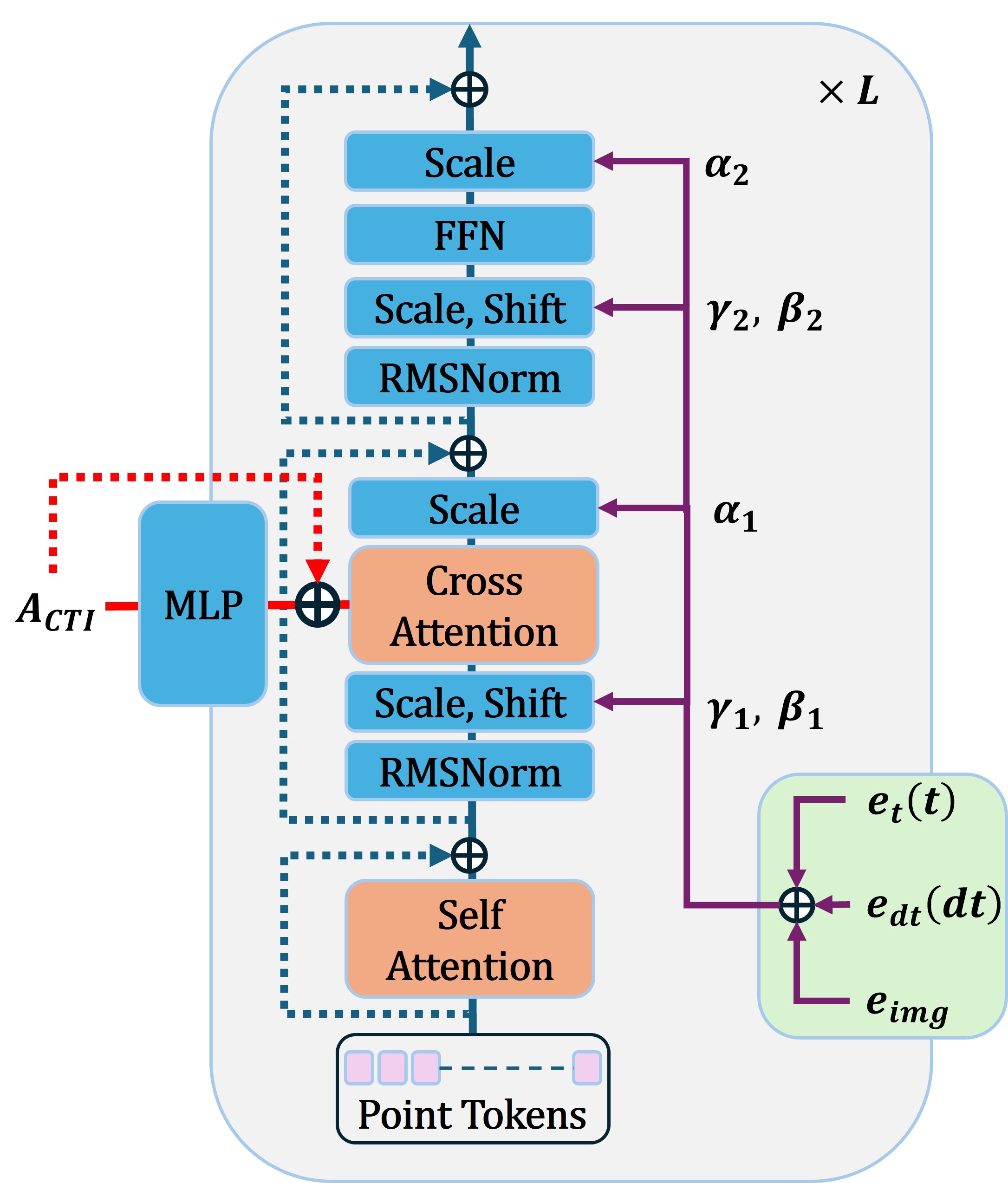}
\caption{
Architecture of a Point-MF DiT block.
Each block consists of Self-Attention, Cross-Attention, and FFN, and is repeated $L$ times.
The conditioning vector composed of $e_t(t)$, $e_{dt}(dt)$, and $e_{\mathrm{img}}$
modulates the block through AdaLN (purple arrows).
The CTI feature $A_{\mathrm{CTI}}$ is injected into the Cross-Attention branch through an MLP (red arrows).
}
\label{fig:dit_block}
\vspace{-1.0\baselineskip}
\end{wrapfigure}

The point-cloud generator of Point-MF consists of $L$ Diffusion Transformer (DiT) blocks, as shown in Fig.~\ref{fig:dit_block}.
Given an input point cloud $x_t\in\mathbb{R}^{N\times 3}$, the 3D coordinates of each point are projected by a linear layer into point tokens with hidden dimension $H$.
Since a point cloud is an unordered set, we do not use positional embeddings that depend on point indices.
Each DiT block consists of three residual branches: Self-Attention, Cross-Attention, and a Feed-Forward Network (FFN).

The point tokens are first updated by Self-Attention to model relationships within the point cloud,
then refer to the image condition through Cross-Attention,
and are finally transformed point-wise by the FFN.
In Cross-Attention, the point tokens are used as queries, while the image context tokens obtained from the DINOv3 patch tokens are used as keys and values.

In Mean Flow, the network predicts the interval-averaged velocity field
$u_\theta(x_t,r,t\mid c)$, rather than the instantaneous velocity field.
Therefore, in addition to the current time $t$, we also provide the interval length $dt=t-r$ as a condition.
The variables $t$ and $dt$ are respectively transformed into
$e_t(t), e_{dt}(dt)\in\mathbb{R}^{H}$ using sinusoidal embeddings followed by two-layer MLPs.
We also project the global image feature obtained from the frozen DINOv3 encoder into
$e_{\mathrm{img}}\in\mathbb{R}^{H}$.
The conditioning vector shared by all DiT blocks is defined as
\begin{align}
I
=
e_t(t)
+
e_{dt}(t-r)
+
e_{\mathrm{img}}.
\label{eq:supp_condition_vector}
\end{align}
From this conditioning vector, modulation parameters are generated for the Self-Attention,
Cross-Attention, and FFN branches in each DiT block.

Let $Y^{(\ell)}\in\mathbb{R}^{N\times H}$ denote the input to the $\ell$-th DiT block.
The update inside the block can be written concisely as
\begin{align}
\tilde{Y}^{(\ell)}
&=
Y^{(\ell)}
+
\mathrm{SA}
\left(
Y^{(\ell)}; I
\right),
\\
\bar{Y}^{(\ell)}
&=
\tilde{Y}^{(\ell)}
+
\mathrm{CA}
\left(
\tilde{Y}^{(\ell)},
Z^{(\ell)}_{\mathrm{ctx}}; I
\right),
\\
Y^{(\ell+1)}
&=
\bar{Y}^{(\ell)}
+
\mathrm{FFN}
\left(
\bar{Y}^{(\ell)}; I
\right),
\label{eq:supp_dit_block}
\end{align}
where $\mathrm{SA}$ and $\mathrm{CA}$ denote Self-Attention and Cross-Attention, respectively.
Here, $Z^{(\ell)}_{\mathrm{ctx}}$ denotes the image context tokens used in the $\ell$-th DiT block.
In each branch, AdaLN-Zero-style modulation generated from the conditioning vector $I$ is applied to RMSNorm-normalized features.

The output of the final DiT block, $Y^{(L)}$, is projected by a linear output layer into a 3D mean velocity for each point.
Therefore, the network output is
\begin{align}
u_\theta(x_t,r,t\mid c)
\in
\mathbb{R}^{N\times 3},
\end{align}
which represents the mean velocity field in point-cloud space over the time interval $[r,t]$.

The image context tokens are constructed from the DINOv3 patch tokens.
Let $Z_{\mathrm{ctx}}\in\mathbb{R}^{M\times D}$ denote the patch-token sequence obtained from DINOv3.
We first project it to the DiT hidden dimension $H$ using the input projection $\phi_{\mathrm{in}}$:
\begin{align}
Z_{\mathrm{base}}
=
\phi_{\mathrm{in}}(Z_{\mathrm{ctx}})
\in
\mathbb{R}^{M\times H}.
\label{eq:supp_zbase}
\end{align}
Next, we apply Multi-Head Self-Attention (MHSA) only once on the condition branch to compute interactions among image tokens:
\begin{align}
A_{\mathrm{CTI}}
=
\mathrm{MHSA}
\left(
\phi_{\mathrm{MHSA}}(Z_{\mathrm{base}})
\right).
\label{eq:supp_acti}
\end{align}
The image context tokens used in the $\ell$-th DiT block are then defined using a layer-specific MLP $\psi_\ell$ as
\begin{align}
Z^{(\ell)}_{\mathrm{ctx}}
=
Z_{\mathrm{base}}
+
\psi_\ell(A_{\mathrm{CTI}}).
\label{eq:supp_cti}
\end{align}
These tokens $Z^{(\ell)}_{\mathrm{ctx}}$ are used as the keys and values in the Cross-Attention of the $\ell$-th block.
The MHSA among image tokens is computed only once on the condition branch, and each DiT block applies only a lightweight layer-specific MLP.

\subsubsection{Training Settings}
Training was conducted on eight NVIDIA RTX A6000 GPUs.
Each training sample consists of a point cloud with $N=2048$ points and an RGB image of size $224\times224$.
We use DINOv3 ViT-B as the image feature extractor and keep its weights frozen during training.
The only pretrained model used in our framework is the image feature extractor;
the point-cloud DiT, Condition-Token Integrator, projection layers, and output layer are trained from scratch.

The point-cloud DiT uses hidden dimension $H=512$, $L=12$ blocks, and $h=8$ attention heads.
The Condition-Token Integrator is applied only once to the DINOv3 patch tokens on the condition branch,
with internal dimension 1024 and 4 attention heads.
Its output projection is zero-initialized.

For time sampling, we use a logit-normal distribution.
The parameters of the underlying normal distribution are $\mu=-0.4$ and $\sigma=1.0$.
During training, we mix the Flow Matching branch and the Mean Flow branch with equal probability.
In the Mean Flow branch, $t$ and $r$ are sampled independently and then reordered by a min--max operation so that $0\le r<t\le 1$.
For the velocity composition in CFG-guided Mean Flow, we use the approximate tangent vector defined in Eq.~\eqref{eq:supp_cfg_vtilde}.
In our experiments, the CFG coefficients are set to $(\omega,\kappa)=(1.0,0.5)$.
The condition dropout rate is set to $0.1$.

The base weight of Denoised Space Anchor (DSA) is set to
$\lambda_{\mathrm{base}}=0.5$.
The stabilization constant used in the mini-batch loss-ratio normalization is
$\delta=10^{-8}$.\begin{wraptable}[7]{r}{0.5\linewidth}
\vspace{-1.2\baselineskip}
\small
\noindent\rule{\linewidth}{0.4pt}\par
\vspace{1pt}
\refstepcounter{algoboxctr}\label{alg:cfg_guided_mf_sampling}
\noindent\textbf{Algorithm \thealgoboxctr. 1-step sampling.}\par
\vspace{-1pt}
\noindent\rule{\linewidth}{0.4pt}

\begin{algobox}
{\ttfamily\scriptsize
\textcolor{algcomment}{\# fn(x\_t, r, t, c): function to predict mean velocity u}\par
\textcolor{algcomment}{\# c: condition input}\par
\textcolor{algcomment}{\# x\_shape: output shape}\par
\vspace{3pt}

e = randn(x\_shape)\par
x = e - fn(e, 0, 1, c)\par
}
\end{algobox}
\vspace{-2pt}
\end{wraptable}
The exponent parameter of the Adaptive L2 loss is set to $p=1$.
We use AdamW for optimization with a learning rate of $1.0\times10^{-4}$.
The first 10,000 steps are used for linear warm-up,
after which the learning rate is kept fixed at $1.0\times10^{-4}$.
The batch size is 128, and the total number of training steps is 120,000.

At inference time, we start from initial noise $\epsilon\sim\mathcal{N}(0,I)$,
fix $(r,t)=(0,1)$, and perform a single Mean Flow update.
Specifically, we generate the point cloud as
\begin{align}
x_0
=
\epsilon
-
u_\theta(\epsilon,0,1\mid c).
\label{eq:supp_impl_one_step}
\end{align}

As shown in Algorithm~\ref{alg:cfg_guided_mf_sampling},
this inference procedure evaluates the network only once, and therefore the NFE is 1.

\section{Core ML Conversion and Mobile Inference}
\label{sec:supp_coreml}

\subsubsection{Core ML Conversion}

\begin{wrapfigure}[27]{r}{0.5\linewidth}
\vspace{-1.2\baselineskip}
\centering
\includegraphics[width=0.95\linewidth]{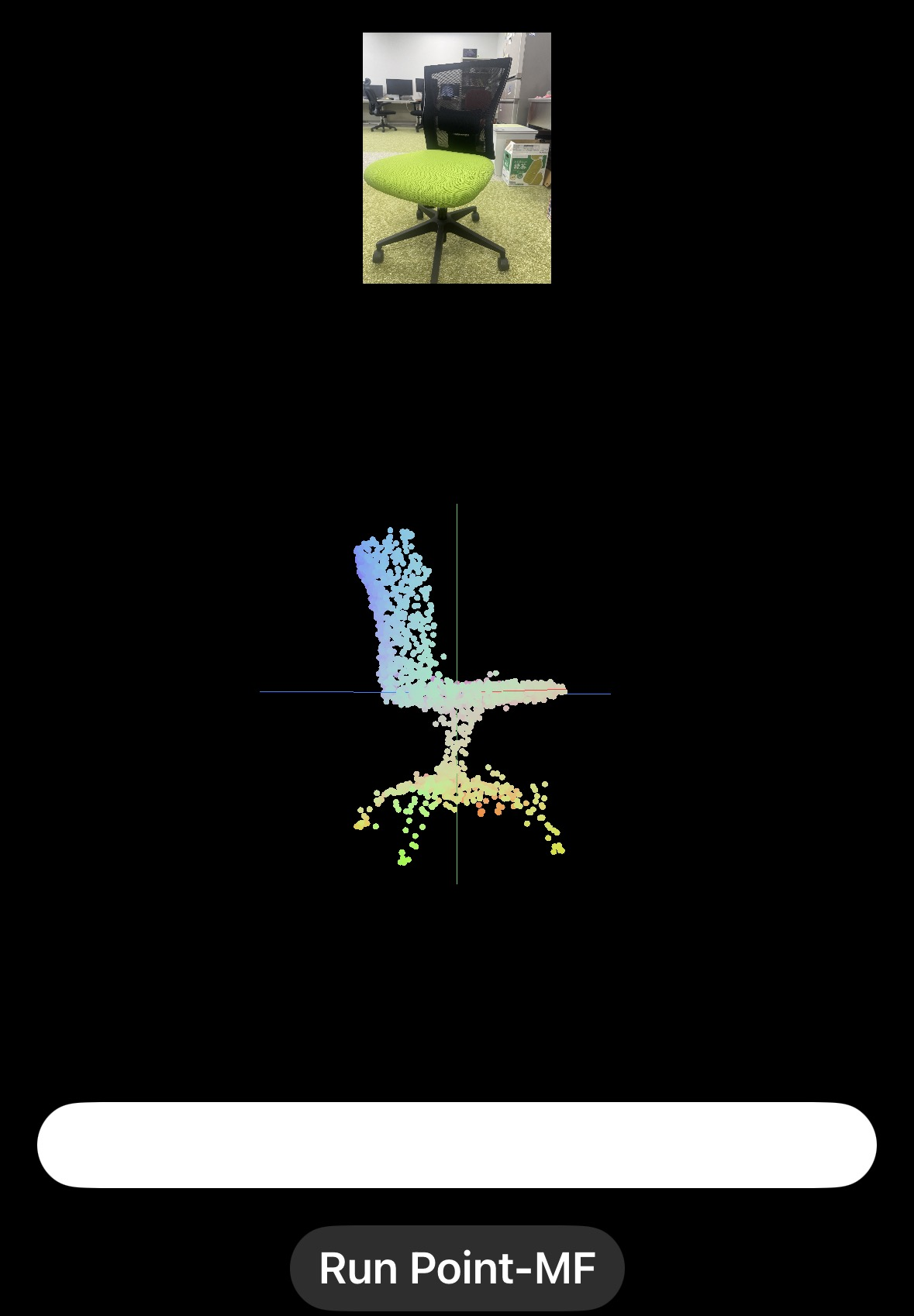}
\caption{
Execution example of the Core ML version of Point-MF running on an iPhone 15 Pro.
When an input RGB image is selected, image-condition features are extracted by DINOv3,
and a 2048-point point cloud is generated and visualized on the device through a 1-step Mean Flow update.
This result demonstrates that the 1-NFE inference of Point-MF can also be executed on a mobile device.
}
\label{fig:ios_demo}
\vspace{-1.0\baselineskip}
\end{wrapfigure}

To verify that the 1-NFE inference of Point-MF can also be executed on a mobile device,
we converted the trained model into the Core ML format and performed inference on an iPhone 15 Pro.
An execution example is shown in Fig.~\ref{fig:ios_demo}.
This evaluation is not intended as a same-environment speed comparison with existing methods,
but rather as a supplementary experiment showing that the 1-step inference of Point-MF works on a real device.

For Core ML conversion, we divided the inference path into two packages.
The first package is the DINOv3 image-conditioning model, which takes an input image
$I_{\mathrm{rgb}}\in\mathbb{R}^{1\times3\times224\times224}$ and outputs
image token features for Cross-Attention,
$Z_{\mathrm{ctx}}\in\mathbb{R}^{1\times200\times768}$,
and a global image feature
$z_{\mathrm{img}}\in\mathbb{R}^{1\times768}$.
The second package is the Point-MF velocity-field model, which takes as input
a noise point cloud $x_1\in\mathbb{R}^{1\times2048\times3}$,
the time pair $(r,t)=(0,1)$, and the DINOv3 features, and outputs the mean velocity field
\begin{align}
\hat{u}
=
u_\theta(x_1,0,1\mid c).
\label{eq:supp_coreml_velocity}
\end{align}

During conversion, we first converted the PyTorch model into a static graph using TorchScript tracing,
and then converted it into the \texttt{mlprogram} format using \texttt{coremltools}.
Because Point-MF does not use Sparse Attention, k-nearest-neighbor operations, or 3D convolutions,
which are often used in 3D generative models, the conversion is relatively straightforward.
The deployment target was set to iOS 16, and the computation precision was set to float16.
In addition, to avoid leaving unsupported attention operations during Core ML conversion,
we replaced Self-Attention and Cross-Attention with explicit attention computations
consisting of query/key/value projections, matrix multiplications, softmax, and value aggregation.
For the DINOv3 model, we also disabled fused attention before conversion.

\subsubsection{1-Step Inference on iOS}
For inference on iOS, we regard $t=1$ as the noise side and $r=0$ as the data side,
and initialize the point cloud as
\begin{align}
x_1=\epsilon,\qquad
\epsilon\sim\mathcal{N}(0,I).
\end{align}
Using the features obtained from the DINOv3 image-conditioning model as the condition $c$,
the Point-MF velocity-field model predicts the mean velocity field $\hat{u}$ in Eq.~\eqref{eq:supp_coreml_velocity}.
We then apply the 1-step Mean Flow update
\begin{align}
\hat{x}_0
=
x_1
-
(t-r)\hat{u}.
\label{eq:supp_coreml_update_general}
\end{align}
Since $(r,t)=(0,1)$, this becomes
\begin{align}
\hat{x}_0
=
\epsilon
-
u_\theta(\epsilon,0,1\mid c).
\label{eq:supp_coreml_update}
\end{align}
This is identical to the 1-step sampling equation used in the main paper.
Since the velocity-field model is evaluated only once on Core ML, the NFE remains 1.

\section{Additional Ablations}
\label{sec:supp_ablation}

\subsubsection{Effect of the DSA Weight $\lambda_{\mathrm{base}}$}

Denoised Space Anchor (DSA) uses the auxiliary-loss weight $\lambda_{\mathrm{base}}$
to geometrically anchor the clean-space point cloud induced by Mean Flow to the ground-truth shape.
When this weight is too small, the geometric regularization from DSA becomes weak,
and outliers or density imbalance caused by the 1-step update tend to remain.
In contrast, when the weight is too large, the point cloud may be pulled too strongly
toward the ground-truth surface.
This can improve CD in some cases, but it may degrade the global point distribution
or suppress fine-grained structures.

To analyze this trade-off without tuning on the real-image test benchmark,
we train Point-MF variants with different values of $\lambda_{\mathrm{base}}$
on ShapeNet-R2N2 and evaluate them on representative ShapeNet-R2N2 categories.
The average values in Table~\ref{tab:compare_cd_emd_shapenet_ab} are weighted by
the number of samples in each category.

\begin{table}[h]
\vspace{-3mm}
  \centering
\caption{
Ablation of the DSA weight $\lambda_{\mathrm{base}}$ on representative ShapeNet-R2N2 categories.
Avg. is weighted by the number of samples per category.
Values may differ from the main experiments because each variant was trained separately.
}
  \vskip -3mm
  \label{tab:compare_cd_emd_shapenet_ab}
      \scriptsize
    \setlength{\tabcolsep}{3pt}
  \begin{tabular*}{\linewidth}{@{\extracolsep{\fill}}lcccccccc}
    \toprule
     & \multicolumn{4}{c}{CD$\times 100$ $\downarrow$} & \multicolumn{4}{c}{EMD$\times 100$ $\downarrow$} \\
    \cmidrule(lr){2-5} \cmidrule(lr){6-9}
    Method & Car & Chair & Air & Avg. & Car & Chair & Air & Avg. \\
    \midrule
    $\lambda_{\mathrm{base}}=1.1$ & \textbf{4.11} & \textbf{4.01} & 2.85 & \textbf{3.79} & 4.13 & 4.31 & 3.09 & 3.97 \\
    $\lambda_{\mathrm{base}}=0.8$ & \underline{4.18} & \underline{4.29} & \underline{2.80} & \underline{3.92} & \underline{3.94} & \underline{4.29} & \textbf{2.98} & \underline{3.86} \\
    $\lambda_{\mathrm{base}}=0.5$ & 4.32 & 4.31 & \textbf{2.76} & 3.97 & \textbf{3.91} & \textbf{4.21} & \underline{3.00} & \textbf{3.82} \\
    $\lambda_{\mathrm{base}}=0.3$ & 5.12 & 5.81 & 3.02 & 4.91 & 4.11 & 5.01 & 3.52 & 4.31 \\
    w/o DSA & 5.21 & 6.16 & 3.71 & 5.23 & 4.89 & 6.08 & 4.12 & 5.16 \\
    \bottomrule
  \end{tabular*}
  \vspace{-5mm}
\end{table}

As shown in Table~\ref{tab:compare_cd_emd_shapenet_ab},
removing DSA clearly degrades both CD and EMD.
The w/o DSA variant obtains weighted average CD and EMD of 5.23 and 5.16, respectively,
whereas all variants with DSA achieve better performance.
This demonstrates that clean-space geometric regularization is effective for suppressing
outliers and density imbalance in 1-step Mean Flow generation.

With $\lambda_{\mathrm{base}}=0.3$, the performance improves over w/o DSA,
but the weighted average CD and EMD are 4.91 and 4.31, respectively,
which are worse than the other DSA settings.
This suggests that the DSA weight is too small to sufficiently anchor the predicted denoised point cloud
to the ground-truth shape.

On the other hand, $\lambda_{\mathrm{base}}=1.1$ achieves the best weighted average CD of 3.79,
but its weighted average EMD is 3.97, which is worse than that of $\lambda_{\mathrm{base}}=0.5$.
Since Chamfer Distance is based on nearest-neighbor distances,
it can improve when points are pulled closer to the ground-truth surface locally,
but it does not necessarily reflect global point distribution or correspondence consistency.
Therefore, choosing the weight based only on CD may lead to less favorable point distributions.

\begin{figure}[h]
  \centering
  \vspace{-3mm}
  \includegraphics[width=1.0\linewidth]{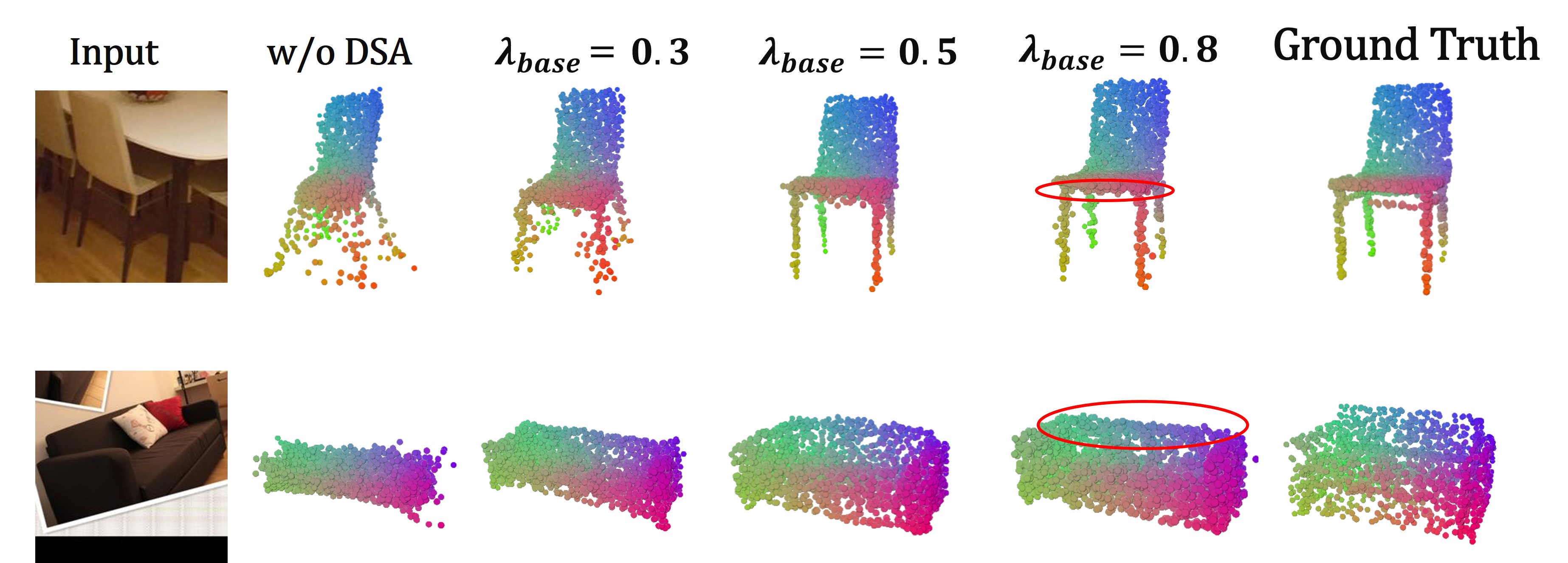}
  \caption{
  Qualitative comparison of the DSA weight $\lambda_{\mathrm{base}}$ on Pix3D.
  When the weight is too small, outliers and density imbalance tend to remain.
  When the weight is too large, some fine-grained structures may be lost,
  as indicated by the red circles.
  }
  \label{fig:ablation_lambda}
  \vspace{-4mm}
\end{figure}

Fig.~\ref{fig:ablation_lambda} shows qualitative Pix3D examples for different values of
$\lambda_{\mathrm{base}}$.
These real-image examples are used only to visualize the qualitative effect of the DSA weight.
When $\lambda_{\mathrm{base}}$ is small, the DSA constraint is weak,
and scattered points or outliers tend to remain in some parts of the point cloud.
In contrast, when $\lambda_{\mathrm{base}}\ge 0.8$, the surface alignment becomes stronger,
but some fine-grained structures of chairs and tables may be lost, as indicated by the red circles.
This qualitative tendency is consistent with the CD--EMD trade-off observed in the ShapeNet-R2N2 ablation.

Based on the ShapeNet-R2N2 ablation, we use $\lambda_{\mathrm{base}}=0.5$ in this work.
Although this setting does not achieve the best weighted average CD,
it achieves the best weighted average EMD of 3.82 and provides a better balance
between local surface alignment and global point-distribution consistency.
The Pix3D qualitative examples further suggest that this setting suppresses outliers
while avoiding excessive smoothing or loss of fine structures.
Therefore, in the main experiments, we use $\lambda_{\mathrm{base}}=0.5$
by considering the ShapeNet-R2N2 CD--EMD trade-off and qualitative shape preservation,
rather than choosing a setting that optimizes CD alone.

\subsubsection{NFE--Quality Trade-off}
\label{sec:supp_nfe_quality}

We examine whether additional sampling steps improve reconstruction quality
and whether DSA remains beneficial beyond one-step sampling.
We evaluate Point-MF and its variant without DSA at 1, 2, 4, and 8 NFEs
on all 13 ShapeNet-R2N2 categories.
For each variant, we use its converged checkpoint without NFE-specific
retraining and use paired initial noise for the comparisons.
CD and EMD are reported as instance-weighted averages over the 13 categories;
category counts are provided in Sec.~\ref{sec:supp_additional_quant}.

\paragraph{Sampling procedures.}
Let $1=t_0>t_1>\cdots>t_M=0$ denote the sampling times,
where $M$ is the NFE, and initialize $x_{t_0}=\epsilon$ with
$\epsilon\sim\mathcal{N}(0,I)$.
ODE sampling composes the predicted interval-averaged updates:
\begin{align}
x_{t_{i+1}}
= x_{t_i}-(t_i-t_{i+1})
u_\theta(x_{t_i},t_{i+1},t_i\mid c).
\label{eq:supp_nfe_ode}
\end{align}
Consistency sampling instead predicts a data-side estimate and re-noises it
to the next sampling time:
\begin{align}
\hat{x}_0^{(i)}
&=x_{t_i}-t_i u_\theta(x_{t_i},0,t_i\mid c),\\
x_{t_{i+1}}
&=(1-t_{i+1})\hat{x}_0^{(i)}+t_{i+1}\xi_i,
\qquad \xi_i\sim\mathcal{N}(0,I).
\label{eq:supp_nfe_consistency}
\end{align}
At the final step, $t_{i+1}=0$, so no noise is added.
Both procedures use the same trained Mean-Flow network and require one
network evaluation per step; they coincide when $M=1$.
Consistency sampling here is a sampling rule, not a separately trained
consistency model.
\begin{figure}[tb]
  \centering
  \includegraphics[width=\linewidth]{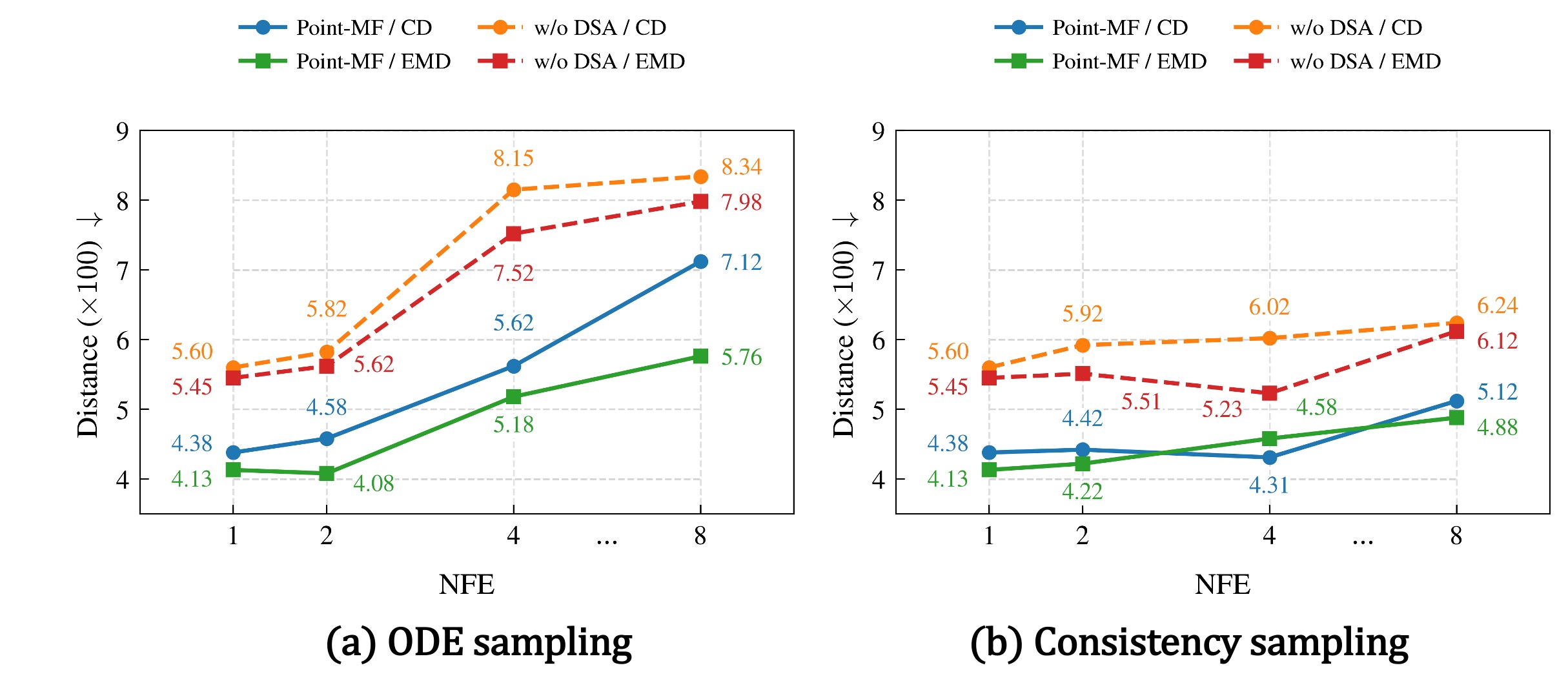}
  \caption{
  NFE--quality comparison on all 13 ShapeNet-R2N2 categories using
  (a) ODE sampling and (b) consistency sampling.
  Each panel compares CD and EMD for Point-MF with and without DSA.
  Results are instance-weighted averages, with distances multiplied by 100;
  lower is better.
  Each variant uses the same converged checkpoint across NFEs without
  NFE-specific retraining, and initial noise is paired.
  Both samplers coincide at 1 NFE.
  }
  \label{fig:supp_nfe_quality}
\end{figure}
\paragraph{Results and interpretation.}
As shown in Fig.~\ref{fig:supp_nfe_quality}, increasing the NFE does not
consistently improve reconstruction quality.
Under ODE sampling, Point-MF changes from CD/EMD of 4.38/4.13 at 1 NFE
to 7.12/5.76 at 8 NFEs.
Consistency sampling mitigates this high-NFE degradation, yielding
5.12/4.88 at 8 NFEs, but does not recover the one-step quality.
DSA improves both metrics at every tested NFE under both samplers.
Under ODE sampling, it reduces the 1-to-8-NFE EMD increase from 2.53
without DSA to 1.63 with DSA, a 35.6\% reduction.
Although some multi-step settings slightly improve an individual metric,
none improves both CD and EMD over the 1-NFE Point-MF result.
Since each additional step requires another network evaluation, these
results support 1 NFE as our default quality--efficiency operating point
for the evaluated checkpoints and sampling procedures.

A related trend is reported for the vanilla MeanFlow baseline in
F2D2~\cite{f2d2}: on CIFAR-10, FID is 2.80, 2.84, 5.14, and 4.34
at 1, 2, 4, and 8 Euler steps, respectively.
Although this concerns image generation rather than point-cloud
reconstruction, it similarly shows that increasing the step count need
not improve a learned Mean-Flow model.

\subsubsection{Output Diversity}
\label{sec:supp_output_diversity}

DSA anchors a generated estimate to a single ground-truth point set,
which could reduce stochastic variation.
We therefore evaluate reconstruction accuracy and output diversity jointly.
We use 60 fixed ShapeNet-R2N2 inputs, with 20 inputs each from airplane,
car, and chair, and generate $K=10$ outputs per input by varying only
the sampling seed.
The Point-MF variants use paired initial noise and one-step sampling.
We compare them with our RGB2point reimplementation~\cite{rgb2point},
PC$^2$ with DDPM sampling~\cite{pc2}, and BDM~\cite{bdm}.

\paragraph{Metrics.}
For an input with generated point sets $X^{(1)},\ldots,X^{(K)}$,
we define Total Mutual Difference (TMD) as
\begin{align}
\mathrm{TMD}_K
=\sum_{k=1}^{K}\frac{1}{K-1}
\sum_{\substack{\ell=1\\\ell\ne k}}^{K}
d_{\mathrm{CD}}\!\left(X^{(k)},X^{(\ell)}\right),
\label{eq:supp_tmd}
\end{align}
where $d_{\mathrm{CD}}$ is L1 Chamfer Distance.
Thus, TMD sums each output's mean distance to the other $K-1$ outputs;
we do not divide this sum by $K$.
For each reconstruction metric $d\in\{d_{\mathrm{CD}},d_{\mathrm{EMD}}\}$,
we report the mean over the $K$ outputs, the minimum error
(Best@$K$), and the standard deviation of the metric across seeds.
Best@$K$ is selected separately for CD and EMD using the ground truth,
so it is an oracle diagnostic rather than an inference-time selection rule.
All statistics are computed per input and then averaged equally over the
60 inputs. This subset differs from the full 13-category evaluation above.
All reported distances, including TMD and seed standard deviations,
are multiplied by 100.

\begin{table}[tb]
  \centering
  \caption{
  Accuracy and diversity on 60 fixed ShapeNet-R2N2 inputs
  (20 per category) with $K=10$ sampling seeds.
  TMD$_{10}$ follows Eq.~\eqref{eq:supp_tmd}.
  Best@10 is the metric-wise oracle, and seed std. measures variation
  in reconstruction error across seeds.
  All statistics are averaged over inputs; distances are multiplied by 100.
  }
  \label{tab:supp_output_diversity}
  \small
  \setlength{\tabcolsep}{4pt}
  \begin{tabular}{@{}lcccc@{}}
    \toprule
    Method & TMD$_{10}$
    & \shortstack{CD $\downarrow$\\Mean/Best@10}
    & \shortstack{EMD $\downarrow$\\Mean/Best@10}
    & \shortstack{Seed std.\\CD/EMD} \\
    \midrule
    RGB2point (reimpl.)
    & 0.000 & 4.39/4.39 & 5.76/5.76 & 0.000/0.000 \\
    PC$^2$ (DDPM)
    & 40.976 & 4.68/4.19 & 4.72/4.02 & 0.446/0.567 \\
    BDM
    & 39.091 & 4.64/4.13 & 4.67/3.92 & 0.300/0.513 \\
    Point-MF w/o DSA
    & 33.121 & 5.28/4.12 & 5.15/3.94 & 0.428/0.441 \\
    Point-MF
    & 30.921 & \textbf{3.91/3.74} & \textbf{3.71/3.41}
    & 0.109/0.197 \\
    \bottomrule
  \end{tabular}
\end{table}

\paragraph{Results and interpretation.}
Table~\ref{tab:supp_output_diversity} shows that Point-MF retains
seed-dependent output variation, unlike deterministic RGB2point
(TMD$_{10}=0$).
DSA reduces TMD from 33.121 to 30.921, retaining 93.4\% of the value
without DSA, while improving mean CD from 5.28 to 3.91 and mean EMD
from 5.15 to 3.71 (25.9\% and 28.0\% reductions).
The seed standard deviations decrease from 0.428/0.441 to 0.109/0.197
for CD/EMD, corresponding to reductions of 74.5\%/55.3\%.
For Point-MF, Best@10 further improves CD/EMD from 3.91/3.71 to
3.74/3.41, indicating that different seeds still yield outputs with
different reconstruction errors.

These results are consistent with DSA suppressing inaccurate
seed-dependent variation while retaining a stochastic mapping,
rather than making the output effectively deterministic on this subset.
TMD alone is not a quality measure: larger point-set differences may
also reflect reconstruction errors, so we interpret it jointly with
accuracy and seed variability.
This analysis measures point-set variation, not exhaustive coverage of
semantically distinct shape completions.

\section{Additional Qualitative Results}
\label{sec:supp_qualitative}

This section presents additional qualitative comparisons on
ShapeNet-R2N2 and Pix3D, including an FCM comparison, a failure case,
and reference comparisons with recent image-to-3D models.
Note that recent image-to-3D models differ from the aligned geometry-only point cloud reconstruction setting studied in this work
in terms of output representation, training data, and evaluation objective.
Therefore, in this section, we treat them as qualitative reference comparisons.

\subsubsection{Additional Comparisons on ShapeNet-R2N2 and Pix3D}
\begin{figure}[p]
  \centering
  \includegraphics[width=1.0\linewidth]{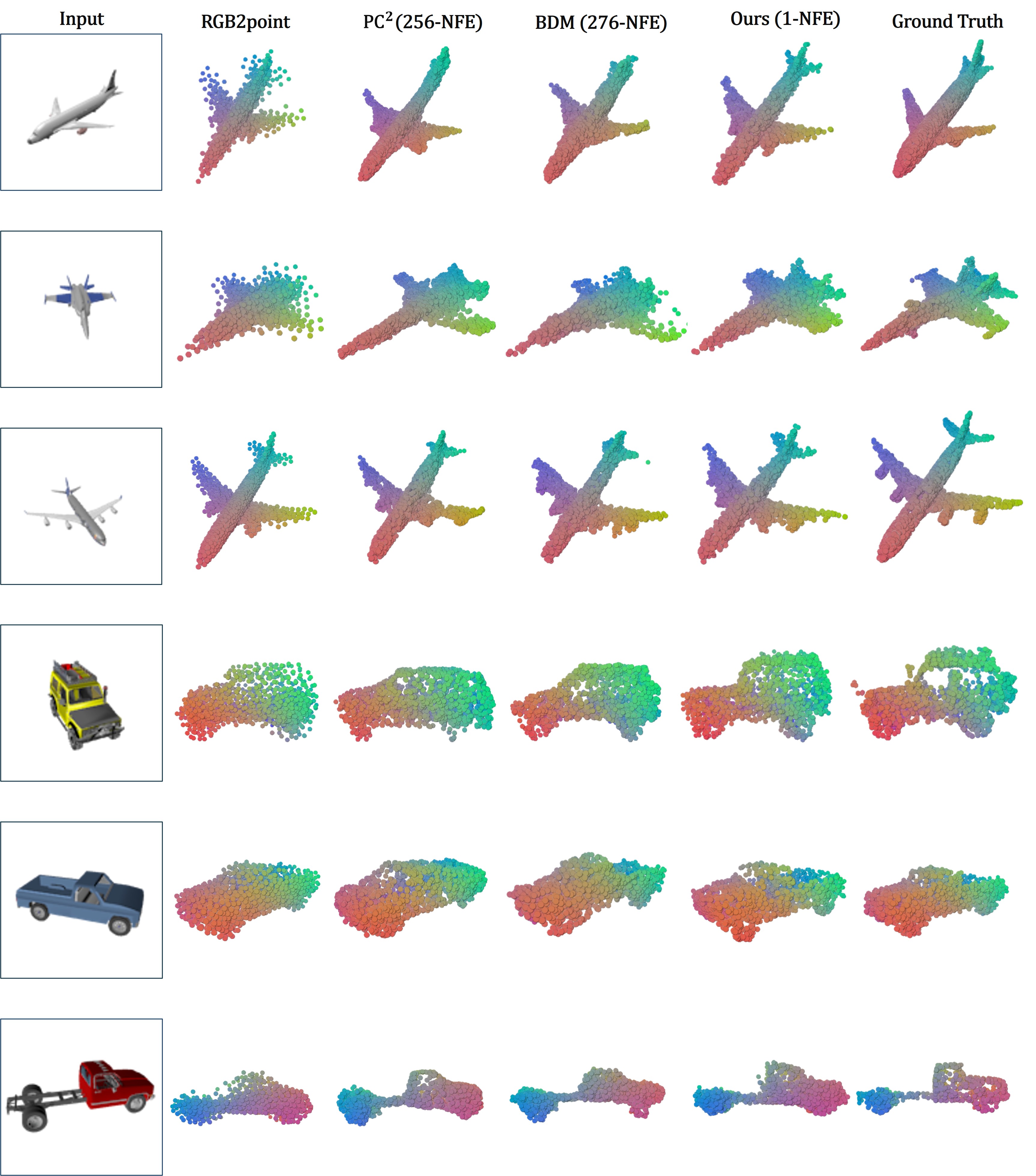}
  \caption{
  Additional qualitative comparison for the airplane and car categories on ShapeNet-R2N2.
  }
  \label{fig:shapenet01}
\end{figure}

\begin{figure}[p]
  \centering
  \includegraphics[width=1.0\linewidth]{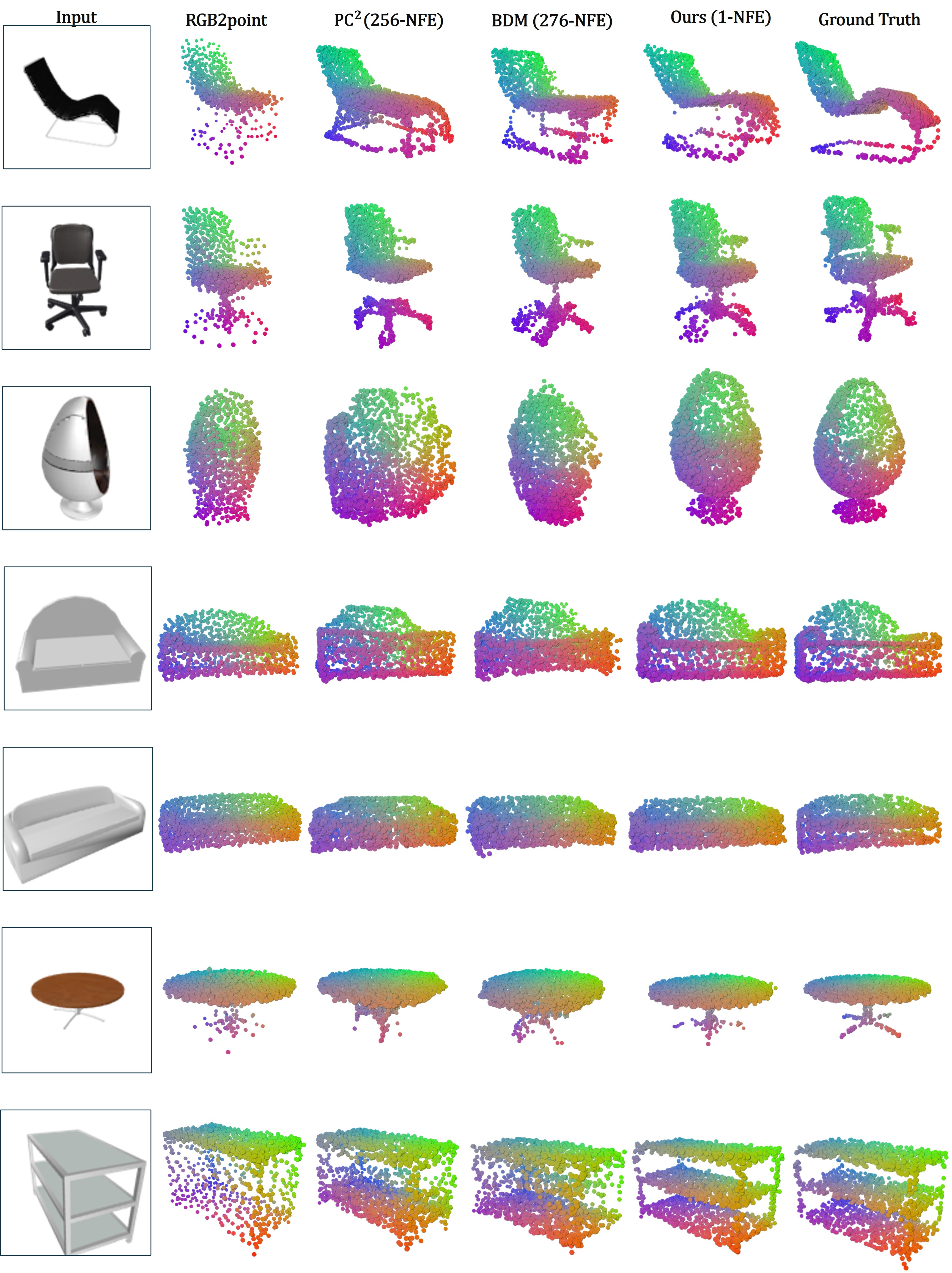}
  \caption{
  Additional qualitative comparison for the chair, sofa, and table categories on ShapeNet-R2N2.
  }
  \label{fig:shapenet02}
\end{figure}

\begin{figure}[p]
  \centering
  \includegraphics[width=1.0\linewidth]{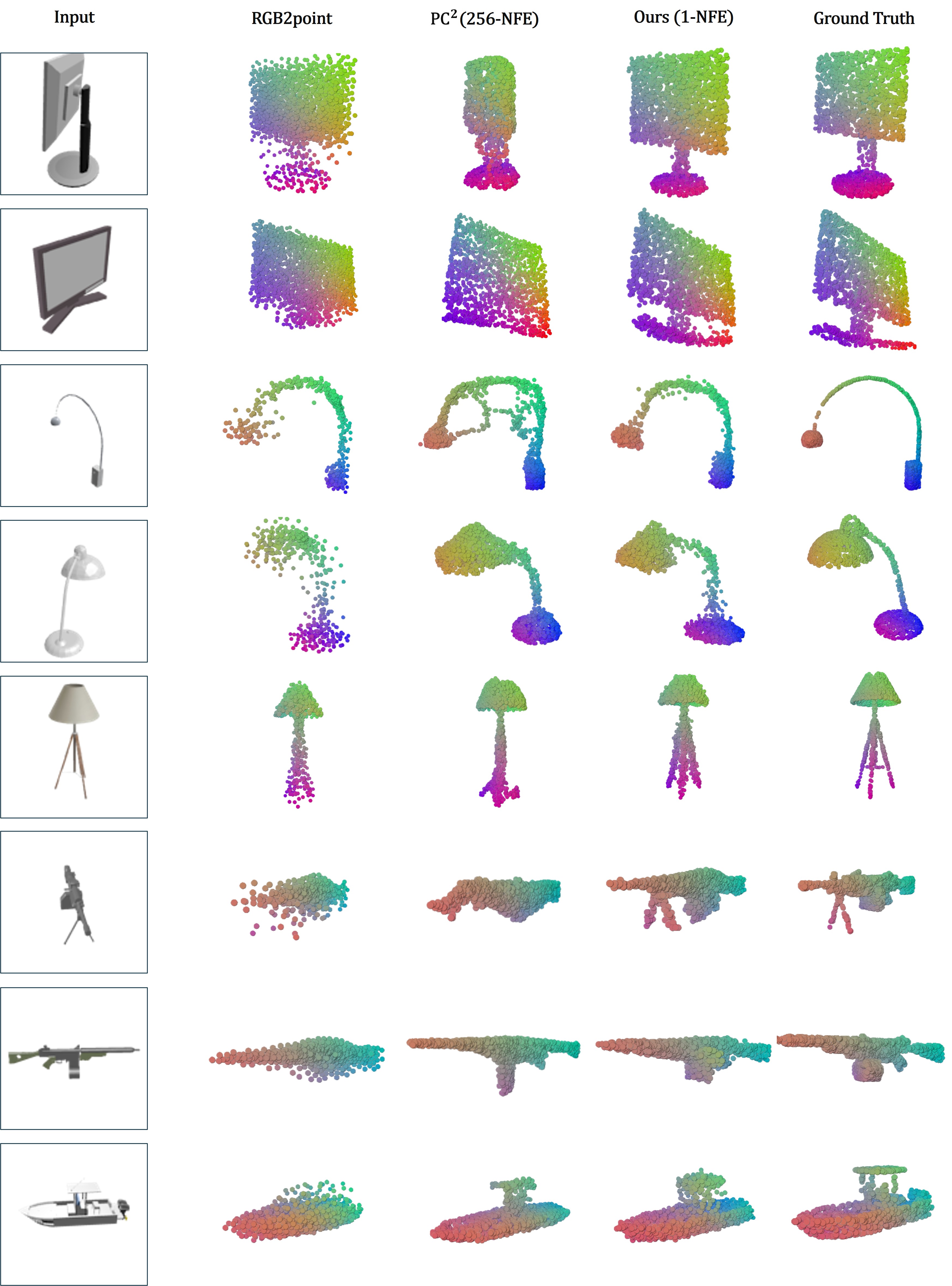}
  \caption{
  Additional qualitative comparison for the display, lamp, and rifle categories on ShapeNet-R2N2.
  }
  \label{fig:shapenet03}
\end{figure}

\begin{figure}[p]
  \centering
  \includegraphics[width=1.0\linewidth]{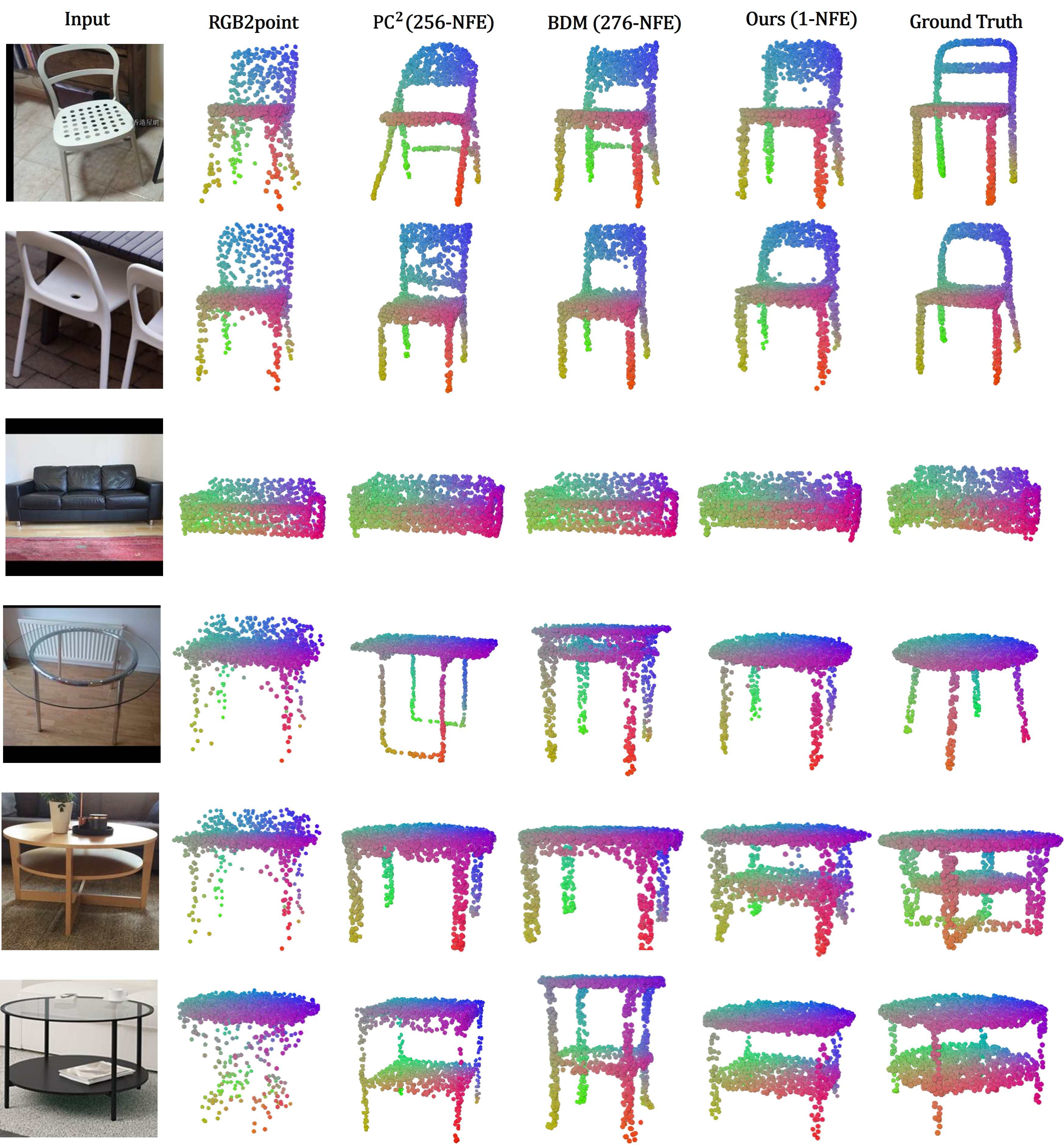}
  \caption{
  Additional qualitative comparison on Pix3D.
  }
  \label{fig:pix3d01}
\end{figure}

Additional qualitative comparisons between the proposed method and existing point cloud reconstruction methods~\cite{rgb2point,pc2,bdm}
on ShapeNet-R2N2 and Pix3D are shown in
Figs.~\ref{fig:shapenet01}--\ref{fig:shapenet03} and Fig.~\ref{fig:pix3d01}, respectively.
These examples supplement the representative cases included in the main paper
and show reconstruction tendencies across more diverse categories and real-image inputs.
Note that BDM requires a category-specific prior model, and therefore its qualitative results are shown only for the major categories.

Overall, Point-MF generates point clouds that preserve the main structures of objects such as chairs, tables, sofas, cars, and airplanes,
despite using 1-NFE inference.
In particular, while RGB2point sometimes produces local point concentration or density imbalance in some regions,
Point-MF tends to maintain a relatively stable point distribution.
Moreover, compared with iterative diffusion-based methods such as PC$^2$ and BDM,
Point-MF can reconstruct the overall object shape with substantially fewer NFEs.
\begin{figure}[t]
  \centering
  \includegraphics[width=1.0\linewidth]{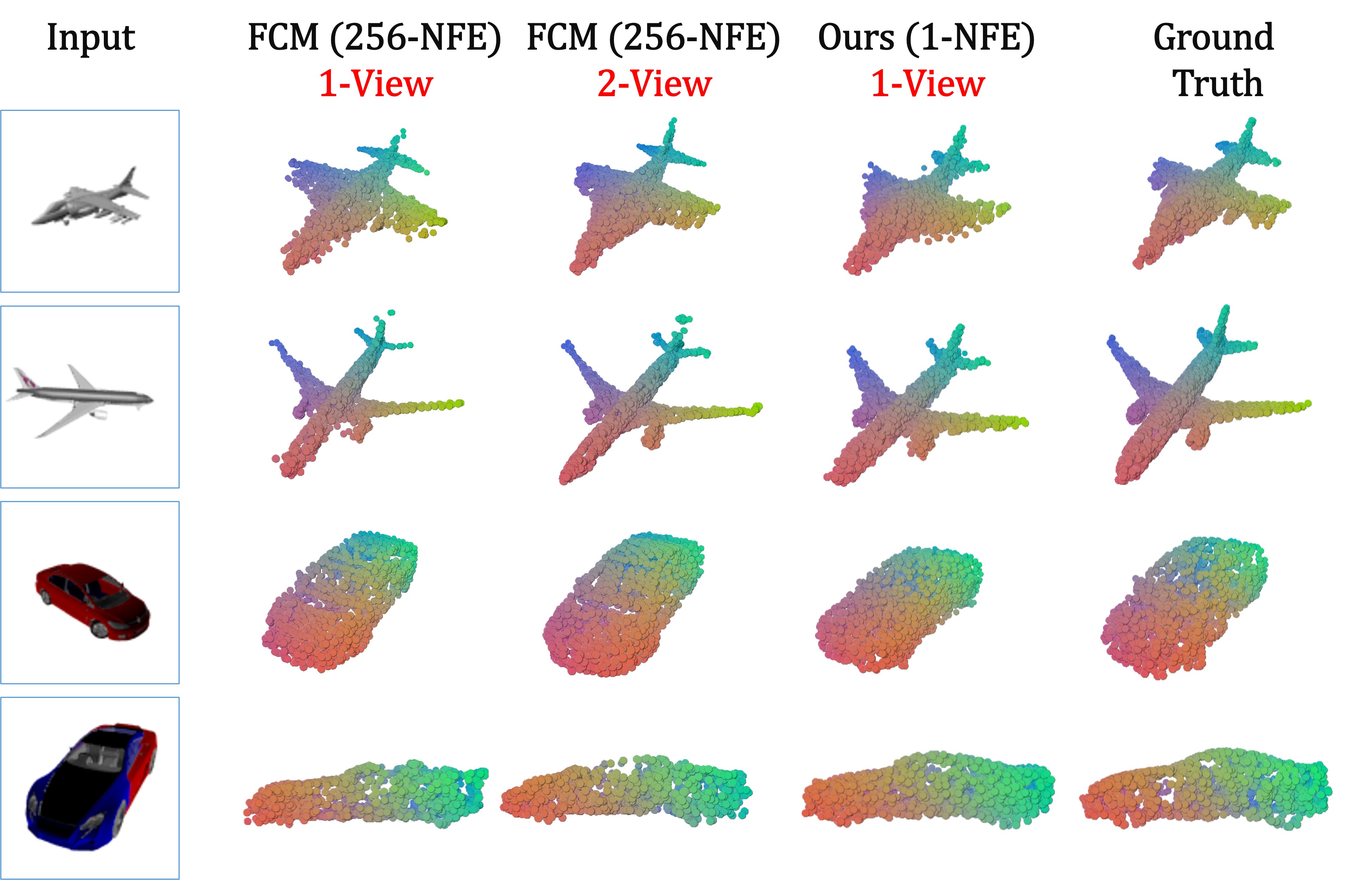}
  \caption{
  Additional qualitative comparison with FCM on ShapeNet-R2N2.
  Since the training code of FCM is not publicly available,
  we report results only for the comparable airplane and car categories.
  }
  \label{fig:fcm_qual}
\end{figure}

\subsubsection{Additional Comparison with FCM}
Fig.~\ref{fig:fcm_qual} shows an additional qualitative comparison with FCM~\cite{fcm}.
Since the training code of FCM is not publicly available, we do not retrain FCM or extend the experiment to all categories.
Instead, we report results only for the comparable airplane and car categories.
This comparison is not intended as a fully matched-condition comparison with FCM,
but rather as a reference comparison against an existing few-view point cloud reconstruction method.

\subsubsection{Failure Case}
\begin{figure}[h]
  \centering
  \vspace{-5mm}
  \includegraphics[width=1.0\linewidth]{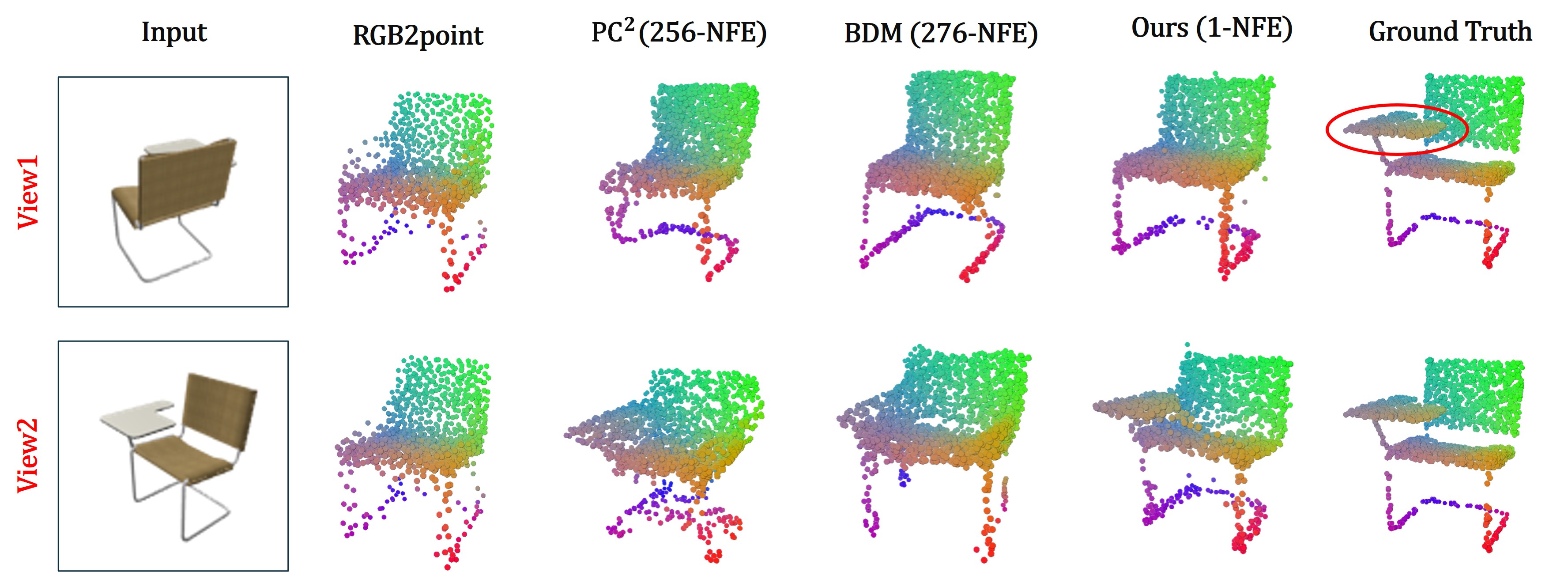}
  \caption{
  Failure case.
  The attached desk highlighted by the red circle in the ground truth
  may not be reconstructed depending on the input viewpoint.
  }
  \label{fig:fault}
  \vspace{-3mm}
\end{figure}
Fig.~\ref{fig:fault} shows a failure case.
In this example, the attached desk highlighted by the red circle in the ground truth
is not reconstructed when it is not visible from the input viewpoint.
This reflects a common limitation of single-image reconstruction methods.
Specifically, single-image reconstruction must infer occluded structures from incomplete visual evidence,
and attached parts or fine details may not be recovered when the input image does not contain sufficient cues.
Furthermore, some existing methods fail to reconstruct this part even when it is visible in the input image.

\subsubsection{Additional Reference Comparisons with Recent Image-to-3D Models}
In the main paper, we presented representative reference comparisons
with ZeroShape~\cite{zeroshape}, SAM3D~\cite{sam3d},
and SPAR3D~\cite{spar3d}.
Here, we add TRELLIS.2~\cite{trellis2} and further examples.
For each model, we sample point clouds from its output and visualize them from the same viewpoint.
Many of these models target output representations that differ from ours,
such as meshes, 3D Gaussians, structured 3D latents, or 3D assets with appearance.
Therefore, this comparison is not a primary comparison for the aligned geometry-only point cloud reconstruction setting targeted by Point-MF,
but rather a reference comparison with recent image-to-3D models.

Specifically, Fig.~\ref{fig:recent_i23d_supp01} shows ShapeNet-R2N2 results
obtained by sampling point clouds from the outputs of recent image-to-3D models,
and Fig.~\ref{fig:recent_i23d_supp02} shows the corresponding Pix3D results.
While Point-MF does not aim to generate textures or high-density meshes,
it directly reconstructs image-aligned object-level geometry as a point cloud with 1-NFE.
In contrast, recent image-to-3D models can generate richer appearance and dense geometry,
but direct comparison based on point-set distances should be interpreted carefully under our evaluation setting
because of differences in scale, orientation, and shape representation.
As shown in Figs.~\ref{fig:recent_i23d_supp01} and~\ref{fig:recent_i23d_supp02},
SAM3D shows strong qualitative results in several open-category cases.
However, its pipeline includes segmentation or foreground-mask preprocessing before 3D generation,
and its output representation and objective differ from our aligned point-set reconstruction setting.
Accordingly, SAM3D, TRELLIS.2, SPAR3D, and ZeroShape should be interpreted as reference comparisons
rather than fully controlled in-domain point-cloud reconstruction baselines.

In addition, such preprocessing and asset-generation pipelines incur substantially higher computational cost
than the one-step inference of Point-MF.
Therefore, these results highlight a trade-off between general-purpose image-to-3D generation
and lightweight aligned point-cloud reconstruction, rather than a direct replacement comparison.

\section{Additional Quantitative Results}
\label{sec:supp_additional_quant}

This section reports quantitative results on all 13 ShapeNet-R2N2 categories.
Some results are obtained from separate experimental runs and may therefore
differ from those reported in the main paper.
We also include SPAR3D, SAM3D, TRELLIS.2, and ZeroShape as
reference comparisons with recent image-to-3D models.

\begin{figure}[p]
  \centering
  \includegraphics[width=1.0\linewidth]{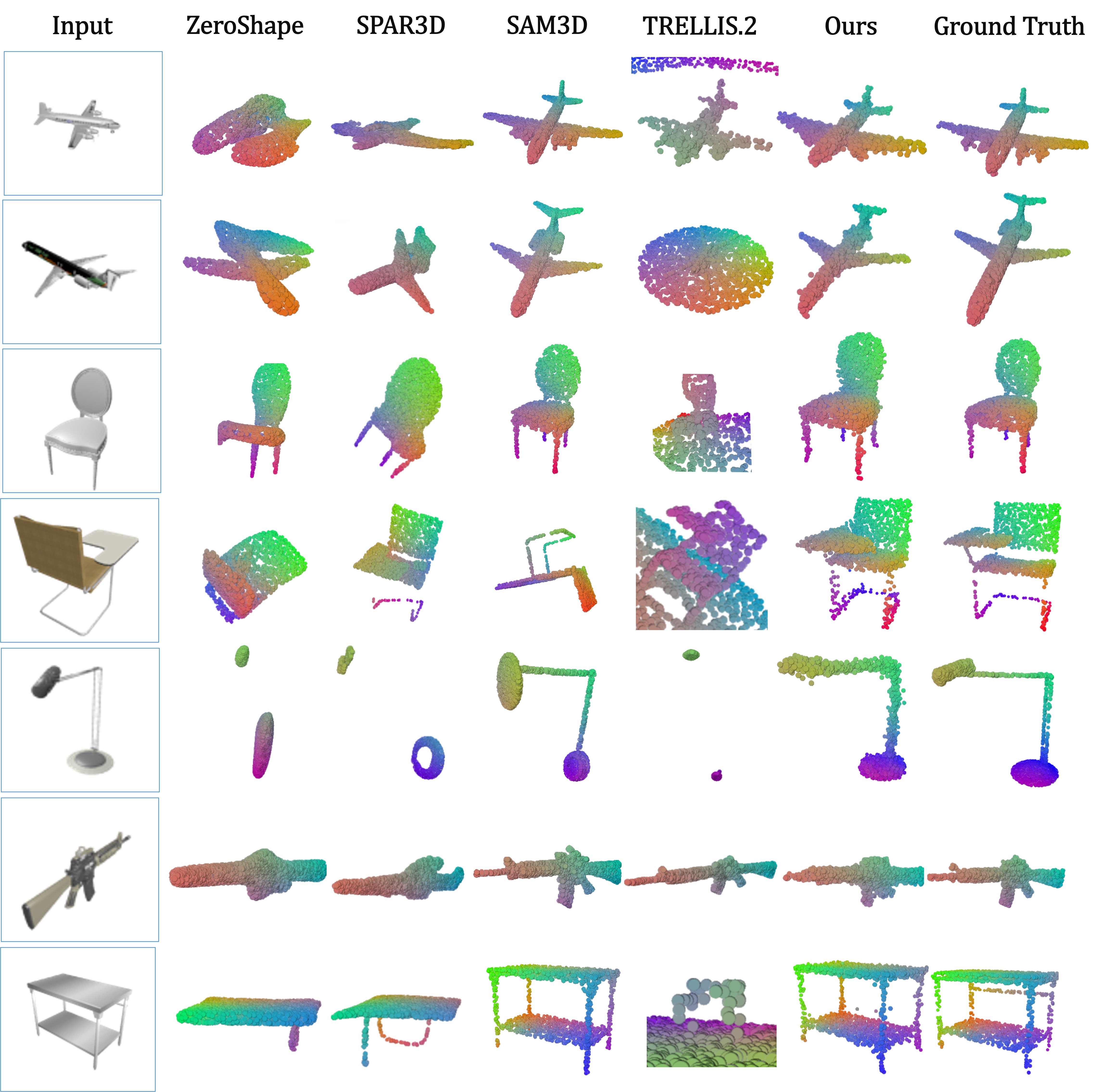}
  \caption{
  Additional reference comparison with recent image-to-3D models on ShapeNet-R2N2.
  Point clouds are sampled from the outputs of each model and visualized.
  }
  \label{fig:recent_i23d_supp01}
\end{figure}

\begin{figure}[p]
  \centering
  \includegraphics[width=1.0\linewidth]{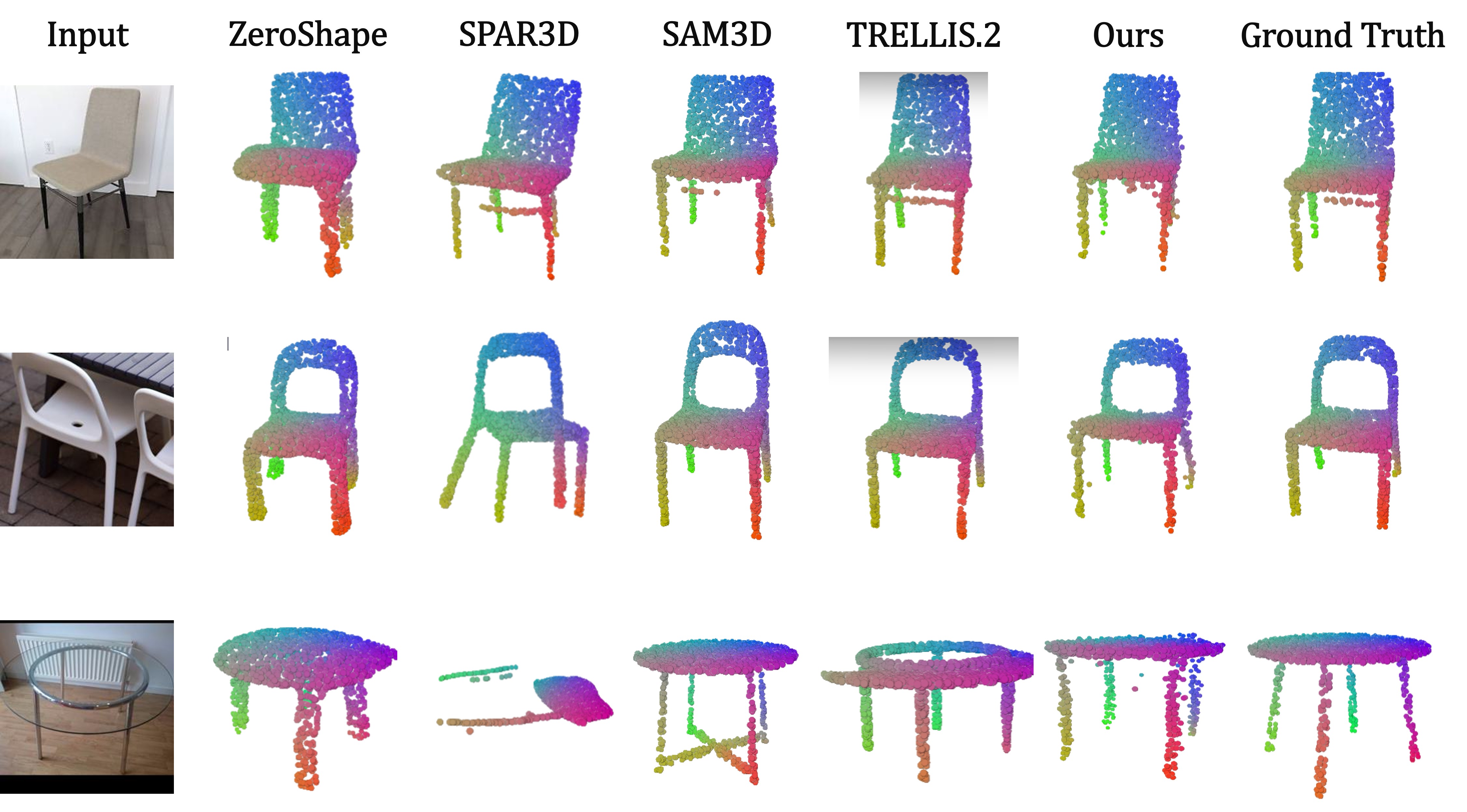}
  \caption{
  Additional reference comparison with recent image-to-3D models on Pix3D.
  Point clouds are sampled from the outputs of each model and visualized.
  }
  \label{fig:recent_i23d_supp02}
\end{figure}

Unlike the aligned geometry-only point cloud reconstruction setting targeted in this work,
these recent image-to-3D models aim at meshes, 3D assets, Gaussian representations,
or segmentation-assisted 3D reconstruction.
Therefore, this comparison is not a fully controlled in-domain comparison based on the same training data,
the same output representation, and the same task setting.
For example, as can be observed in the qualitative results in
Figs.~\ref{fig:recent_i23d_supp01} and~\ref{fig:recent_i23d_supp02},
many outputs differ in rotation or shape.
In this work, we sample point clouds from the outputs of each method,
apply scale and rotation alignment to the ground truth,
and evaluate them using the same CD, EMD, and F-score@1\% code.
Thus, these results should be interpreted as reference values obtained by converting recent image-to-3D models
to our aligned point-set protocol.

Tables~\ref{tab:supp_shapenet_cd}, \ref{tab:supp_shapenet_emd}, and
\ref{tab:supp_shapenet_fscore} show the results for all 13 categories.
Here, $N_c$ denotes the number of instances in category $c$, and
Avg.\ denotes the $N_c$-weighted mean over these categories.
Class avg.\ denotes the unweighted mean over the same categories.
R2P denotes our RGB2point reimplementation, and SP3D denotes SPAR3D.
MF(ViT) denotes a Point-MF variant in which the image feature extractor is replaced with an
ImageNet-pretrained ViT-B~\cite{VisionTransformer} instead of DINOv3.
MF denotes the proposed method, Point-MF.
For F-score@1\%, checkmarks in Table~\ref{tab:supp_shapenet_fscore}
indicate RepKPU upsampling from 2048 to 8192 points;
unmarked Point-MF results use direct 8192-point generation.

\begin{table*}[tb]
\centering
\caption{
CD results on all 13 ShapeNet-R2N2 categories. Lower CD is better.
R2P, SP3D, and Ours (ViT) denote RGB2Point, SPAR3D, and our model with a ViT-B encoder, respectively.
$\dagger$ denotes recent image-to-3D models not directly trained on ShapeNet-R2N2;
their results are reference comparisons obtained by sampling point clouds from model outputs.
}
\label{tab:supp_shapenet_cd}
\vspace{-2mm}
\footnotesize
\setlength{\tabcolsep}{4.2pt}
\renewcommand{\arraystretch}{0.86}
\resizebox{\textwidth}{!}{%
\begin{tabular}{
l r
c c
>{\columncolor{blue!6}}c
>{\columncolor{blue!6}}c
>{\columncolor{blue!6}}c
>{\columncolor{blue!6}}c
c c
}
\toprule
Cat. & $N_c$
& R2P~\cite{rgb2point}
& PC$^2$~\cite{pc2}
& SP3D$^\dagger$~\cite{spar3d}
& SAM3D$^\dagger$~\cite{sam3d}
& TRELLIS.2$^\dagger$~\cite{trellis2}
& ZeroShape$^\dagger$~\cite{zeroshape}
& Ours(ViT)
& Ours \\
\midrule
airplane   & 810  & \textbf{2.70} & 3.37 & 6.34 & 2.97 & 9.38 & 8.64 & \underline{2.72} & 2.76 \\
bench      & 364  & 4.00 & 4.87 & 7.65 & 4.14 & 16.31 & 10.89 & \underline{3.90} & \textbf{3.88} \\
cabinet    & 315  & \textbf{5.36} & 6.25 & 8.34 & 5.59 & 16.93 & 7.68 & 5.71 & \underline{5.53} \\
car        & 1501 & \underline{4.22} & 4.96 & 10.08 & 4.56 & 14.61 & 7.85 & \textbf{4.12} & \underline{4.22} \\
chair      & 1357 & 5.43 & 5.60 & 7.58 & 4.46 & 14.40 & 13.67 & \underline{4.34} & \textbf{4.29} \\
display    & 220  & \textbf{4.74} & 6.07 & 7.66 & 4.91 & 10.42 & 10.19 & 5.02 & \underline{4.84} \\
lamp       & 465  & 6.02 & \underline{5.67} & 7.18 & \textbf{5.03} & 18.94 & 11.84 & 5.92 & 5.98 \\
speaker    & 325  & \underline{6.30} & 7.07 & 9.51 & \textbf{6.04} & 17.94 & 8.02 & 6.53 & 6.46 \\
rifle      & 475  & \underline{2.66} & 3.09 & 6.44 & 3.26 & 7.93 & 5.15 & 2.68 & \textbf{2.55} \\
sofa       & 635  & \underline{5.19} & 5.56 & 8.40 & 5.28 & 15.15 & 9.53 & 5.23 & \textbf{5.14} \\
table      & 1703 & 4.76 & 5.24 & 6.72 & \textbf{4.50} & 15.00 & 10.35 & \underline{4.61} & 4.65 \\
phone      & 211  & \textbf{3.93} & 5.14 & 6.70 & 5.50 & 10.00 & 7.18 & 4.33 & \underline{4.26} \\
watercraft & 389  & 4.43 & 4.60 & 9.11 & 4.83 & 9.74 & 7.31 & \underline{4.11} & \textbf{4.10} \\
\midrule
Avg. & -- & 4.58 & 5.08 & 7.85 & 4.51 & 13.92 & 9.64 & \underline{4.382} & \textbf{4.376} \\
Class avg. & -- & 4.60 & 5.19 & 7.82 & 4.70 & 13.60 & 9.10 & \underline{4.56} & \textbf{4.51} \\
\bottomrule
\end{tabular}%
}
\vspace{-6mm}
\end{table*}

Table~\ref{tab:supp_shapenet_cd} shows the full 13-category Chamfer Distance results.
Point-MF achieves performance that is nearly tied with the best average CD.
RGB2point obtains strong CD on several categories, which is likely because it is trained with a CD-based objective.
However, since CD is based on nearest-neighbor distances,
it does not necessarily fully reflect the global point distribution or coverage.
Therefore, in this work, we evaluate EMD and F-score@1\% in addition to CD.

\begin{table*}[tb]
\centering
\caption{
EMD results on all 13 ShapeNet-R2N2 categories. Lower EMD is better.
R2P, SP3D, and Ours (ViT) denote RGB2Point, SPAR3D, and our model with a ViT-B encoder, respectively.
$\dagger$ denotes recent image-to-3D models not directly trained on ShapeNet-R2N2;
their results are reference comparisons obtained by sampling point clouds from model outputs.
}
\label{tab:supp_shapenet_emd}
\vspace{-2mm}
\footnotesize
\setlength{\tabcolsep}{4.2pt}
\renewcommand{\arraystretch}{0.86}
\resizebox{\textwidth}{!}{%
\begin{tabular}{
l r
c c
>{\columncolor{blue!6}}c
>{\columncolor{blue!6}}c
>{\columncolor{blue!6}}c
>{\columncolor{blue!6}}c
c c
}
\toprule
Cat. & $N_c$
& R2P~\cite{rgb2point}
& PC$^2$~\cite{pc2}
& SP3D$^\dagger$~\cite{spar3d}
& SAM3D$^\dagger$~\cite{sam3d}
& TRELLIS.2$^\dagger$~\cite{trellis2}
& ZeroShape$^\dagger$~\cite{zeroshape}
& Ours(ViT)
& Ours \\
\midrule
airplane   & 810  & 6.86 & 4.28 & 7.89 & 4.05 & 11.30 & 10.13 & \underline{3.12} & \textbf{3.04} \\
bench      & 364  & 5.30 & 4.82 & 8.49 & 4.68 & 18.80 & 12.36 & \underline{3.69} & \textbf{3.56} \\
cabinet    & 315  & 7.37 & 5.47 & 8.34 & 5.55 & 18.64 & 7.70 & \underline{4.62} & \textbf{4.52} \\
car        & 1501 & 5.63 & 4.70 & 9.22 & 4.85 & 17.27 & 8.79 & \underline{4.02} & \textbf{3.82} \\
chair      & 1357 & 9.53 & 5.37 & 8.14 & 5.07 & 15.55 & 14.17 & \underline{4.31} & \textbf{4.29} \\
display    & 220  & 5.71 & 5.52 & 8.12 & 5.61 & 12.35 & 11.35 & \underline{4.82} & \textbf{4.21} \\
lamp       & 465  & 15.07 & \textbf{6.49} & 9.28 & 7.83 & 20.92 & 14.31 & 6.62 & \underline{6.52} \\
speaker    & 325  & 8.77 & 6.19 & 9.57 & 6.15 & 18.90 & 8.08 & \underline{5.55} & \textbf{5.42} \\
rifle      & 475  & 7.83 & 4.45 & 9.39 & 5.97 & 10.06 & 7.81 & \underline{3.56} & \textbf{3.37} \\
sofa       & 635  & 6.94 & 4.97 & 8.45 & 5.21 & 17.95 & 10.39 & \underline{4.49} & \textbf{4.29} \\
table      & 1703 & 7.91 & 5.09 & 7.25 & 5.04 & 16.76 & 11.23 & \textbf{4.21} & \underline{4.24} \\
phone      & 211  & 4.03 & 4.50 & 6.71 & 6.10 & 11.57 & 8.65 & \underline{3.91} & \textbf{3.50} \\
watercraft & 389  & 5.57 & 4.63 & 9.79 & 6.00 & 11.47 & 8.86 & \underline{3.81} & \textbf{3.79} \\
\midrule
Avg. & -- & 7.63 & 5.04 & 8.39 & 5.26 & 15.86 & 10.72 & \underline{4.24} & \textbf{4.13} \\
Class avg. & -- & 7.42 & 5.11 & 8.51 & 5.55 & 15.50 & 10.29 & \underline{4.36} & \textbf{4.20} \\
\bottomrule
\end{tabular}%
}
\vspace{-6mm}
\end{table*}

Table~\ref{tab:supp_shapenet_emd} shows the full 13-category EMD results.
Point-MF achieves the best average EMD among the compared methods.
This result indicates that Point-MF not only places points near the ground-truth surface,
but also maintains a stable distribution over the entire point cloud.
Compared with MF(ViT), Point-MF using DINOv3 also improves the average EMD,
suggesting that DINOv3 features are effective as conditioning information for preserving global point-cloud consistency.

\begin{table*}[tb]
\centering
\caption{
F-score@1\% results on all 13 ShapeNet-R2N2 categories. Higher is better.
$\checkmark$~in the Upsampling row indicates RepKPU upsampling.
LF, R2P, SP3D, and Ours (ViT) denote LegoFormer, RGB2Point, SPAR3D, and our ViT-B encoder variant, respectively.
$\dagger$ marks reference comparisons from point clouds sampled from recent image-to-3D model outputs.
}
\label{tab:supp_shapenet_fscore}
\vspace{-2mm}
\footnotesize
\setlength{\tabcolsep}{3.6pt}
\renewcommand{\arraystretch}{0.86}
\resizebox{\textwidth}{!}{%
\begin{tabular}{
l r
c c c
>{\columncolor{blue!6}}c
>{\columncolor{blue!6}}c
>{\columncolor{blue!6}}c
>{\columncolor{blue!6}}c
c c c
}
\toprule
Cat. & $N_c$
& LF~\cite{lego}
& R2P~\cite{rgb2point}
& PC$^2$~\cite{pc2}
& SP3D$^\dagger$~\cite{spar3d}
& SAM3D$^\dagger$~\cite{sam3d}
& TRELLIS.2$^\dagger$~\cite{trellis2}
& ZeroShape$^\dagger$~\cite{zeroshape}
& Ours(ViT)
& Ours
& Ours \\
\midrule
Upsampling
& --
& --
& --
& $\checkmark$
& --
& --
& --
& --
& $\checkmark$
& --
& $\checkmark$ \\
\midrule
airplane   & 810  & 0.215 & 0.583 & 0.473 & 0.252 & 0.555 & 0.183 & 0.248 & \underline{0.602} & 0.591 & \textbf{0.603} \\
bench      & 364  & 0.241 & 0.406 & 0.326 & 0.216 & \underline{0.488} & 0.163 & 0.180 & 0.482 & 0.473 & \textbf{0.494} \\
cabinet    & 315  & \textbf{0.308} & 0.162 & 0.151 & 0.138 & \underline{0.214} & 0.122 & 0.167 & 0.183 & 0.199 & 0.200 \\
car        & 1501 & 0.220 & 0.339 & 0.301 & 0.108 & \textbf{0.349} & 0.112 & 0.172 & 0.336 & 0.338 & \underline{0.342} \\
chair      & 1357 & 0.217 & 0.195 & 0.202 & 0.192 & \textbf{0.312} & 0.149 & 0.088 & \underline{0.252} & 0.243 & \underline{0.252} \\
display    & 220  & 0.261 & 0.235 & 0.190 & 0.165 & \textbf{0.295} & 0.215 & 0.173 & 0.272 & \underline{0.274} & 0.263 \\
lamp       & 465  & 0.220 & 0.211 & 0.252 & 0.285 & \textbf{0.399} & 0.110 & 0.166 & 0.280 & 0.276 & \underline{0.290} \\
speaker    & 325  & \textbf{0.286} & 0.113 & 0.122 & 0.116 & \underline{0.182} & 0.093 & 0.144 & 0.124 & 0.144 & 0.142 \\
rifle      & 475  & 0.364 & 0.674 & 0.548 & 0.304 & 0.670 & 0.360 & 0.460 & \underline{0.712} & 0.691 & \textbf{0.722} \\
sofa       & 635  & \textbf{0.260} & 0.194 & 0.177 & 0.139 & \textbf{0.260} & 0.128 & 0.144 & 0.226 & 0.238 & \underline{0.240} \\
table      & 1703 & 0.305 & 0.268 & 0.238 & 0.207 & \textbf{0.369} & 0.184 & 0.124 & 0.312 & 0.309 & \underline{0.316} \\
phone      & 211  & \textbf{0.575} & 0.372 & 0.274 & 0.243 & 0.386 & 0.230 & 0.329 & 0.381 & \underline{0.401} & 0.389 \\
watercraft & 389  & 0.283 & 0.406 & 0.348 & 0.157 & 0.422 & 0.210 & 0.275 & \underline{0.442} & 0.438 & \textbf{0.457} \\
\midrule
Avg. & -- & 0.265 & 0.316 & 0.279 & 0.188 & \textbf{0.377} & 0.164 & 0.177 & 0.349 & 0.347 & \underline{0.356} \\
Class avg. & -- & 0.289 & 0.320 & 0.277 & 0.194 & \textbf{0.377} & 0.174 & 0.205 & 0.354 & 0.355 & \underline{0.362} \\
\bottomrule
\end{tabular}%
}
\vspace{-6mm}
\end{table*}

Table~\ref{tab:supp_shapenet_fscore} shows the F-score@1\% results for all 13 categories.
Among the in-domain point-cloud reconstruction baselines, Point-MF achieves the best average F-score@1\%.
This trend is broadly consistent with the EMD results in Table~\ref{tab:supp_shapenet_emd},
where Point-MF achieves the best average EMD among the evaluated methods.
These results indicate that Point-MF can stably preserve the overall object shape and point-cloud coverage
under the aligned geometry-only point-cloud reconstruction protocol.

Ours (ViT) also shows competitive performance, but the DINOv3-based Point-MF improves the average EMD
and performs well on categories such as bench, display, rifle, phone, and watercraft.
This suggests that DINOv3 features are effective for improving global point-cloud consistency,
although the gains are not solely attributable to the image encoder.

SPAR3D, SAM3D, TRELLIS.2, and ZeroShape are treated as reference comparisons
under our aligned point-set protocol.
Although these methods can generate richer appearance and dense geometry,
their output representations, training data, and evaluation objectives differ from those of this work.
In addition, methods such as SAM3D require an additional segmentation model or foreground mask extraction
before 3D generation.
Therefore, these results should be interpreted not as fully controlled in-domain comparisons
with point-cloud reconstruction methods such as RGB2point and PC$^2$,
but as reference values obtained by converting recent image-to-3D model outputs to our evaluation protocol.
SAM3D obtains a higher average F-score under this converted reference protocol,
whereas Point-MF achieves competitive F-score while requiring only 1-NFE.

These complete 13-category results reinforce the trends shown in the main paper:
Point-MF reduces diffusion/flow-based point-cloud reconstruction to 1-NFE
and achieves the best average EMD, while obtaining competitive F-score among
in-domain point-cloud reconstruction baselines under the aligned ShapeNet-R2N2 protocol.

\clearpage

\clearpage

\end{document}